\documentclass{article}

\PassOptionsToPackage{numbers, compress}{natbib}

\usepackage[preprint]{neurips_2019}

\usepackage{selectp}

\usepackage[utf8]{inputenc} 
\usepackage[T1]{fontenc}    
\usepackage{hyperref}       
\usepackage{url}            
\usepackage{booktabs}       
\usepackage{amsfonts}       
\usepackage{nicefrac}       
\usepackage{microtype}      

\usepackage{graphicx}
\usepackage{amsmath,amssymb} 

\usepackage{ulem}

\usepackage{dsfont}        
\newcommand{\R}{\mathds{R}}
\newcommand{\Z}{\mathds{Z}}
\newcommand{\hpu}{HPU-Net}
\newcommand{\pu}{sPU-Net}
\usepackage{siunitx}
\setcitestyle{square}
\setcitestyle{citesep={,}}
\usepackage[linesnumbered]{algorithm2e}
\hypersetup{colorlinks,linkcolor={blue},citecolor={blue},urlcolor={magenta}}  
\renewcommand{\vec}[1]{\boldsymbol{\mathbf{#1}}}
\usepackage{tablefootnote}

\graphicspath{{./lo_res_figures/}}

\title{A Hierarchical Probabilistic U-Net \\ for Modeling Multi-Scale Ambiguities}

\author{%
\textbf{Simon A. A. Kohl}$^{1,2}$, 
\textbf{Bernardino Romera-Paredes}$^1$, 
\textbf{Klaus H. Maier-Hein}$^3$, \\
\textbf{Danilo Jimenez Rezende}$^1$, 
\textbf{S. M. Ali Eslami}$^1$, 
\textbf{Pushmeet Kohli}$^1$, 
\textbf{Andrew Zisserman}$^1$, \\
\textrm{and}~\textbf{Olaf Ronneberger}$^1$\\ \\
$^1$ \textrm{DeepMind, London, UK}\\
$^2$ \textrm{Karlsruhe Institute of Technology, Karlsruhe, Germany}\\
$^3$ \textrm{Division of Medical Image Computing, German Cancer Research Center, Heidelberg, Germany}\\
\texttt{\{simonkohl,brp,danilor,aeslami,pushmeet,zisserman,olafr\}@google.com} \\
\texttt{k.maier-hein@dkfz.de}
}

\begin{document}

\maketitle

\begin{abstract}
  Medical imaging only indirectly measures the molecular identity of the tissue within each voxel, which often produces only ambiguous image evidence for target measures of interest, like semantic segmentation. This diversity and the variations of plausible interpretations are often specific to given image regions and may thus manifest on various scales, spanning all the way from the pixel to the image level. In order to learn a flexible distribution that can account for multiple scales of variations, we propose the Hierarchical Probabilistic U-Net, a segmentation network with a conditional variational auto-encoder (cVAE) that uses a hierarchical latent space decomposition. We show that this model formulation enables sampling and reconstruction of segmenations with high fidelity, i.e. with finely resolved detail, while providing the flexibility to learn complex structured distributions across scales. We demonstrate these abilities on the task of segmenting ambiguous medical scans as well as on instance segmentation of neurobiological and natural images. Our model automatically separates independent factors across scales, an inductive bias that we deem beneficial in structured output prediction tasks beyond segmentation.
\end{abstract}

\section{Introduction}
\begin{figure}[!h]
\centering
\includegraphics[width=\textwidth]{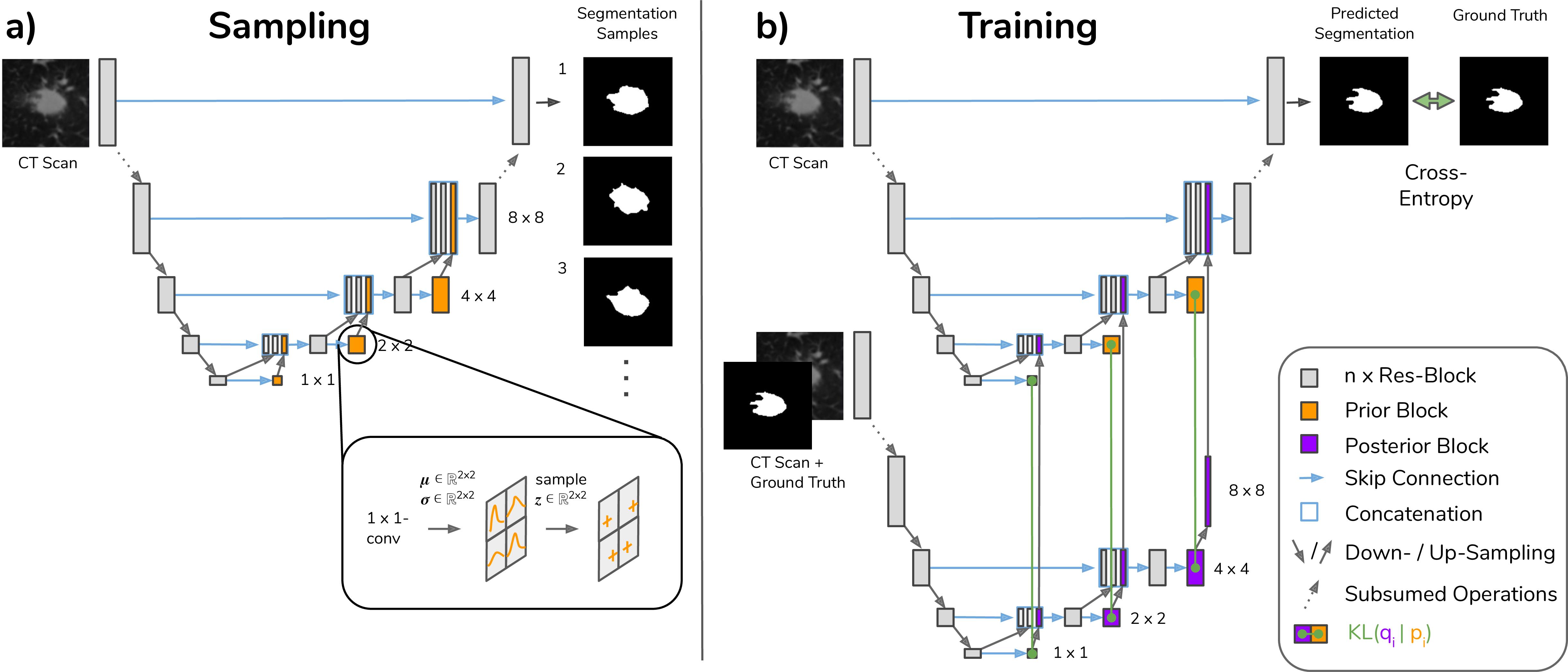}
\caption{The Hierarchical Probabilistic U-Net. The model is based on a U-Net and adds a hierarchy of spatially arranged Gaussian distributions that is interleaved with the U-Net's decoder. (\textbf{a}) Sampling process: For each iteration of the network latents $\vec{z}_i$ at scale $i$ (slim orange blocks) are successively sampled from the prior when going up the hierarchy towards increasing resolutions. (\textbf{b}) Training process illustrated for one training example: During training samples $\vec{z}_i$ from the posterior (slim purple blocks) are injected into the U-Net's decoder and used to reconstruct a given segmentation. Green connections: loss functions. For more details see \autoref{sec:architecture} and \autoref{appendix:training_and_architecture}.}
\label{fig:architecture}
\end{figure}

In real world applications the recorded measurements are often not sufficient to predict a single outcome. Instead, we often have a large manifold of potential interpretations, and we can only use the measurements to narrow down this manifold to a smaller manifold. The modelling of the remaining ambiguities and uncertainties is often challenging, especially in high-dimensional outputs, like segmentation maps.
This particularly arises in the fine-grained distinctions that need to be made in medical images. For example a lesion in a CT scan can have 
identical shape and gray values independently of whether it consists of cancer or healthy cells in the real world (see \autoref{fig:architecture}a). In the same way the true shape of the structure might not be resolvable due to fuzzy borders, occlusions and noise. Similar ambiguities also appear in natural images, e.g. think of cats and dogs lying under a sofa with only parts of their fur being visible. An uncertainty-aware segmentation algorithm can assign a 50\% cancer / 50\% non-cancer probability (or 50\% cat / 50\% dog probability) to each pixel \cite{kendall2015bayesian, kendall2017uncertainties}, but depending on the downstream task this pixel-wise probability is not sufficient, because it only provides the marginals of the high-dimensional probability distribution. For example, in a potential clinical application a clinician with access to additional non-imaging data can select the correct segmentation, or a number of segmentation hypotheses can be presented to a subsequent classification network to assign a diagnosis to each possible interpretation of the medical scan \cite{de2018clinically}. Several algorithms have been proposed that provide samples from the output distribution (here: consistent segmentation maps instead of pixel-wise samples). They are based on ensembles \cite{lakshminarayanan2017simple}, networks with multiple heads \cite{batra2012diverse, lee2015m, lee2016stochastic, rupprecht2017learning}, or image-conditional generative models \cite{isola2017image, zhu2017unpaired, zhu2017toward, liu2017unsupervised, park2019semantic} such as cVAEs \cite{vae1, vae2, vae3, sohn2015learning, esser2018variational}, as in the Probabilistic U-Net \cite{kohl2018probabilistic}. These type of approaches work well for a single object in the image or for other global variations (like different segmentation styles, e.g. more narrow or more inclusive outlining), but do not scale to images containing multiple objects with uncorrelated variations. There exist several approaches which use hierarchical latents to produce rich probability distributions \cite{gregor2016towards, sonderby2016ladder, kingma1606improving, salimans2017pixelcnn++, menick2018generating, maaloe2019biva, de2019hierarchical}, but this concept has not yet been used in the context of segmentation or image-to-image translation.

Here we propose a `Hierarchical Probabilistic U-Net' (the \hpu) that overcomes these issues. Similar to the existing Probabilistic U-Net \cite{kohl2018probabilistic} it combines a segmentation U-Net \cite{Ronneberger2015} with a cVAE. Instead of global latent variables we use a coarse-to-fine hierarchy of latent variable maps (see \autoref{fig:architecture}) that are injected into the synthesis path of the U-Net at the corresponding resolutions.

Our main contributions are: (1) A generative model for semantic segmentation able to learn complex-structured conditional distributions equipped with a latent space that scales with image size. This results in (2) Compared to prior art, strongly improved fidelity to fine structure in the models' samples and reconstructions. (3) Improved modelling of distributions over segmentations including independently varying scales and locations, as demonstrated in its ability to generate instance segmentations. (4) Automatic learning of factors of variations across space and scale.

We demonstrate the improved quality of the segmentations on a lung lesion segmentation task. Furthermore we show the ability of the model to learn highly complex probability distributions, by presenting an instance segmentation task, where we ask the model to label (`colorize') each instance consistently with a random instance id. We test this ability on neuronal structures in EM images as well as on car instances in natural images. Finally we show that the model is also capable of predicting consistent segmentations with corresponding uncertainties in a blacked out region of the image. In a medical application this could be used to predict disease progression by applying a 4D version of the proposed network to time series (3 spatial axes and 1 time axis), where the blacked out part corresponds to the unknown future development of the disease. The model will be open-sourced at \url{https://github.com/project_repo}\footnote{The url will be updated once available.}.

\section{Network Architecture and Learning Objective}
\label{sec:architecture}

The standard Probabilistic U-Net (abbreviated as `\pu') models segmentation ambiguities using a low-dimensional, image global latent vector, that is sampled from a separate `prior net' and is combined with U-Net features by means of a shallow network of $1 \times 1$-convolutions \cite{kohl2018probabilistic}. As we show below, this image-global latent space heavily constrains the granularity at which the output space can be modelled. While our proposed architecture also combines a U-Net with a cVAE, it instead employs a hierarchical latent space that resides in the U-Net's decoder. A hierarchical decomposition yields a much more flexible generative model that can further easily model top-down dependencies. E.g. the global part can model the patient's genetic predisposition for a certain disease, while the local parts can model indiscernible tissue types, or fuzzy borders at different scales. The spatial arrangement of the latent variables further enables the network to easily model local independent variations (like multiple lesions). Due to the fully-convolutional architecture, it can also generalize from few to many lesions at arbitrary locations. Beside these fundamental extensions, we additionally removed the separate prior net and instead use 
U-Net internal features to predict the parameters of the prior distributions (as in \cite{esser2018variational}), which results in parameter and run-time savings. For the network to employ the full hierarchy, we further found it crucial to minimize obstructions between latent scales by introducing (pre-activated) res-blocks \cite{he2016identity} (as discussed in \autoref{appendix:training_and_architecture} and in line with \cite{kingma1606improving, maaloe2019biva}).

\textbf{Sampling} The architecture's main feature is its highly flexible parameterization of the conditional prior that it employs. This prior is composed of a) a deterministic feature extractor that computes features at spatial resolutions up to scale $L$ (counted with ascending resolution) for the given input image $X$ and b) a cascade of distributions interleaved with the U-Net's decoder, that allows to hierarchically sample latents. In a conventional U-Net, the U-Net decoder's features of every resolution are up-sampled and then concatenated with the features of the U-Net's encoder from the respective resolution above \cite{Ronneberger2015}. In our proposed architecture there is one additional step at each scale of the latent hierarchy: Conditioned on the decoder features of each scale $i \leq L$, we sample a spatial grid of latents $\vec{z}_i$ and concatenate it with the input decoder features, before the usual up-sampling and concatenation with encoder features from above takes place, see \autoref{fig:architecture}a. The latents of each scale $i$ thus depend on the input image $X$ and on all latents of scales $i' < i$ that have already been sampled lower in the hierarchy, which we collectively denote as $\vec{z}_{<i} := (\vec{z}_{i-1},...,\vec{z}_0)$. At each scale with spatial dimensions $H_i \times W_i$ the model uses conditional Gaussian distributions with mean $\vec{\mu}^{\textrm{prior}}_i \in \R^{H_i \times W_i}$ and variance $\vec{\sigma}^{\textrm{prior}}_i \in \R^{H_i \times W_i}$. The means and variances are predicted by $1 \times 1$-convolutions for each spatial position of that scale. Sampling from the corresponding Gaussian distribution results in the spatial latents $\vec{z}_i \in \R^{H_i \times W_i}$:
\vspace{-1mm}
\begin{equation}
\label{eq:prior}
    \vec{z}_i \sim {\cal{N}}\big(\vec{\mu}^{\textrm{prior}}_i(\vec{z}_{<i}, X), \vec{\sigma}^{\textrm{prior}}_i(\vec{z}_{<i}, X)\big) =: p(\vec{z}_i| \vec{z}_{<i}, X).
\end{equation}
Our experiments did not benefit from going beyond scalar latents at each spatial location, which however is a choice that one might want to make depending on the application. The hierarchical (ancestral) sampling results in a joint distribution for the prior that decomposes as follows:
\vspace{-1mm}
\begin{equation}
    P(\vec{z}_0, ..., \vec{z}_L|X) = p(\vec{z}_L|\vec{z}_{<L}, X)\cdot ... \cdot p(\vec{z}_0|X).
\end{equation}
Every run of the network yields a segmentation hypothesis $Y' = S(X, \vec{z})$ for the given image (where $\vec{z} = (\vec{z}_L, ..., \vec{z}_0)$ and $S$ stands for the segmentation network), which is illustrated in \autoref{fig:architecture}a. Note that only the U-Net's decoder (including the hierarchical sampling) needs to be rerun to produce the next segmentation samples for the same image. The number of latent scales $L$ is a hyper-parameter and typically chosen smaller than the full number of scales of the U-Net; our models use $L = 3$ (4 scales).

\textbf{Training} As is standard practice for VAEs, the training procedure aims at maximizing the so-called evidence lower bound (ELBO) on the likelihood $p(Y|X)$, where in our case $Y$ is a segmentation and $X$ is an image. This requires to model a variational posterior $Q(.|X,Y)$ that depends on both $X$ and $Y$. The structure matches with that of the prior:
\vspace{-1mm}
\begin{align}
\label{eq:posterior}
    \vec{z}_i \sim {\cal{N}}\big(\vec{\mu}^{\textrm{post}}_i(\vec{z}_{<i}, X, Y), \vec{\sigma}^{\textrm{post}}_i(\vec{z}_{<i}, X, Y)\big) =: q(\vec{z}_i| \vec{z}_{<i}, X, Y), \\
    Q(\vec{z}_0, ..., \vec{z}_L|X, Y) = q(\vec{z}_L|\vec{z}_{<L}, X, Y)\cdot ... \cdot q(\vec{z}_0|X, Y).
\end{align}
The posterior $Q$ is modeled in form of a separate network with the same hierarchical topology in which for each scale $i \leq L$, we compute conditional Gaussian distributions with mean $\vec{\mu}^{\textrm{post}}_i \in \R^{H_i \times W_i}$ and variance $\vec{\sigma}^{\textrm{post}}_i \in \R^{H_i \times W_i}$. During training, samples $\vec{z} \sim Q$ are fed into the U-Net's decoder (as illustrated in the bottom half of \autoref{fig:architecture}b) with the aim of learning to reconstruct the given input segmentation $Y$. The reconstruction objective (${\cal{L}}_{\textrm{rec}}$) is formulated as a cross-entropy loss between the prediction $Y'$ and the target $Y$ (below formulated as a pixel-wise categorical distribution $P_c$). Additionally there is a Kullback-Leibler divergence $D_{\mathrm{KL}}(Q||P) = \mathbb{E}_{\vec{z} \sim Q} \left[\mathrm{log}\, Q- \mathrm{log}\, P\right]$, that assimilates $P$ and $Q$ (more details in \autoref{appendix:KL}). Our ELBO objective with a relative weighting factor $\beta$ thus amounts to
\begin{equation}
    \label{eq:elbo}
    {\cal L}_{\textrm{ELBO}} = \mathbb{E}_{\vec{z} \sim Q} \big[-\mathrm{log}\, P_{\textrm{c}}(Y|S(X,\vec{z}))\big] +\beta \cdot \sum_{i=0}^L \mathbb{E}_{\vec{z}_{<i} \sim Q} D_{\mathrm{KL}}\big(q_i(\vec{z}_i|\vec{z}_{<i},X,Y)||p_i(\vec{z}_i|\vec{z}_{<i},X)\big).
\end{equation}
Minimizing ${\cal L}_{\mathrm{ELBO}}$ leads to sub-optimally converged priors in our experiments. For this reason we make use of the recently proposed $GECO$-objective \cite{rezende2018taming} that adds in a constraint on the reconstruction term and thus dynamically balances it with the KL terms from above:
\begin{equation}
    \label{eq:geco}
    {\cal L}_{\textrm{GECO}} = \lambda \cdot \Big(\mathbb{E}_{\vec{z} \sim Q} \big[-\mathrm{log}\, P_{\textrm{c}}(Y|S(X,\vec{z}))\big] - \kappa \Big) +\sum_{i=0}^L \mathbb{E}_{\vec{z}_{<i} \sim Q} D_{\mathrm{KL}}\big(q_i(\vec{z}_i|\vec{z}_{<i},X,Y)||p_i(\vec{z}_i|\vec{z}_{<i},X)\big),
\end{equation}
where $\kappa$ is chosen as the desired reconstruction loss value and the Lagrange multiplier $\lambda$ is updated as a function of the exponential moving average of the reconstruction constraint. This formulation initially puts high pressure on the reconstruction and once the desired $\kappa$ is reached it increasingly moves the pressure over on the KL-term. For more details we refer to \autoref{appendix:training_and_architecture} and the literature \cite{rezende2018taming}.

We additionally perform an online hard-negative mining, specifically, we only back-propagate the gradient of the $k$th percentile of the worst pixels of the batch \cite{wu2016bridging}, ${\cal{L}}_{\textrm{rec}} \rightarrow {\tt{top\_k\_mask}}\left({\cal{L}}_{\textrm{rec}}\right)$. We chose $k = 0.02$ (the worst $2\%$ pixels) in all experiments of the \hpu~and stochastically pick the $k$th percentile \cite{nikolov2018deep} (we sample from a Gumbel-Softmax distribution \cite{jang2016categorical} over ${\cal{L}}_{\textrm{rec}}$ per pixel).

\section{Results}

The \pu~has established significant performance advantages over other approaches in segmenting ambiguous images \cite{kohl2018probabilistic}. With this work we aim at improving on the flexibility of the \pu~to model complex output interdependencies as well as segmentation fidelity. To this end we report results for a segmentation task of CT scans showing potential lung abnormalities annotated by four expert graders (examples are shown in \autoref{fig:lidc}). We further consider the task of segmenting individual instances, i.e. inferring a latent id for each object in an image, 
and use it to assess the models' ability to capture correlated pixel-uncertainty. We use the EM dataset of the SNEMI3D challenge (published in \cite{kasthuri2015saturated}), which contains instance segmentations of neuronal cells (examples are shown in \autoref{fig:snemi3d}) and further probe our model's performance on the segmentation of car instances on natural street scenes from Cityscapes (see \autoref{fig:cityscapes_instances}). For training details in the respective tasks we refer to \autoref{appendix:training_and_architecture}.

\begin{table}[htbp]
  \caption{Test set results. Mean and standard deviation are calculated from results of 10 random model initializations and 1000 bootstraps with replacement. Data splits are defined in \autoref{appendix:data}.}
  \label{tab:results}
  \centering
  \scalebox{0.75}{
  \begin{tabular}{llcc}
    \toprule
    Dataset & Metric &  Probabilistic U-Net &  Hierarchical\\
    &&(re-implementation)& Probabilistic U-Net\\
    \midrule
    \textbf{a) LIDC}&IoU$_{\textrm{rec}}$ &  0.75 $\pm$~0.04 & \textbf{0.97} $\pm$~\textbf{0.00} \\
    &Hungarian-matched IoU & 0.50 $\pm$~0.03 & \textbf{0.53} $\pm$~\textbf{0.01} \\
    &Hungarian-matched IoU (subset B) & 0.37 $\pm$~0.07 & \textbf{0.47} $\pm$~\textbf{0.01} \\
    \midrule
    \textbf{b) SNEMI3D} & IoU$_{\textrm{rec}}$ & 0.13 $\pm$~0.03 & \textbf{0.60} $\pm$~\textbf{0.00}\\
    &Rand Error & 0.52 $\pm$~0.10 & \textbf{0.06} $\pm$~\textbf{0.00}\\
    \midrule
    \textbf{c) Cityscapes}&IoU$_{\textrm{rec}}$ & - & 0.62 \\
    \textbf{\hspace{2.5mm} Car Instances} & Rand Error & - & 0.13 \\
    \bottomrule
  \end{tabular}}
\end{table}

\subsection{Performance Measures}
\label{sec:measures}

We are interested in assessing how well the conditional distribution produced by the respective generative model and the given ground-truth distribution agree, which we measure in terms of the intersection-over-union (IoU) based \textit{Generalized Energy Distance} ($\textrm{GED}^2$) and the \textit{Hungarian-matched IoU}. Furthermore, we would like to measure an upper-bound on the fidelity of the models' samples, i.e. how accurately the models are able to produce fine segmentation structure and detail, for which we employ the \textit{reconstruction IoU},  $\textrm{IoU}_{\textrm{rec}}(Y, Y')$ where $Y' = S(X, \vec{\mu}^{\textrm{post}}(X,Y))$. To assess the instance segmentation performance we employ the metric used in the SNEMI3D challenge, the \textit{Rand Error}. For more details and definitions of all metrics, we refer to \autoref{appendix:metrics}. 

\subsection{LIDC: Segmentation of Ambiguous Lung Scans}
The LIDC-IDRI dataset \cite{armato2015, armato2011lung, clark2013cancer} contains 1018 lung CT scans from 1010 lung patients with manual lesion segmentations from four experts. We process and use the data in the exact same way as \cite{kohl2018probabilistic}, see \autoref{appendix:data}, i.e. the models are fed lesion-centered 2D crops of size $128 \times 128$ for which at least one grader segmented a lesion, resulting in 8882 images in the training set, 1996 images in the validation set and 1992 images in the test set.

The LIDC results are reported in \autoref{tab:results}a. The \hpu~performs better in terms of the Hungarian-matched IoU (and in terms of $\textrm{GED}^2 = 0.27 \pm 0.01$), while showing a largely improved reconstruction fidelity, that amounts to a near perfect posterior reconstruction of $\textrm{IoU}_{\textrm{rec}} = 0.97$. Retraining the \pu~with an identical training set-up as in \cite{kohl2018probabilistic}, we obtain an unsatisfactorily low value of $0.75$ for the foreground-restricted reconstruction IoU ($\textrm{IoU}_{\textrm{rec}}$) and recapture \cite{kohl2018probabilistic}'s $\textrm{GED}^2$ of 0.29 (re-implementation: $\textrm{GED}^2 = 0.32 \pm 0.03$). We additionally evaluate the models on the test subset of samples for which all 4 graders agree on the presence of an abnormality (`subset B', see \autoref{appendix:data}), exposing the \hpu's significantly improved ability to capture shape variations (see also \autoref{appendix:ged_subset_b}).

\begin{figure}[tp]
\centering
\includegraphics[width=0.8\textwidth]{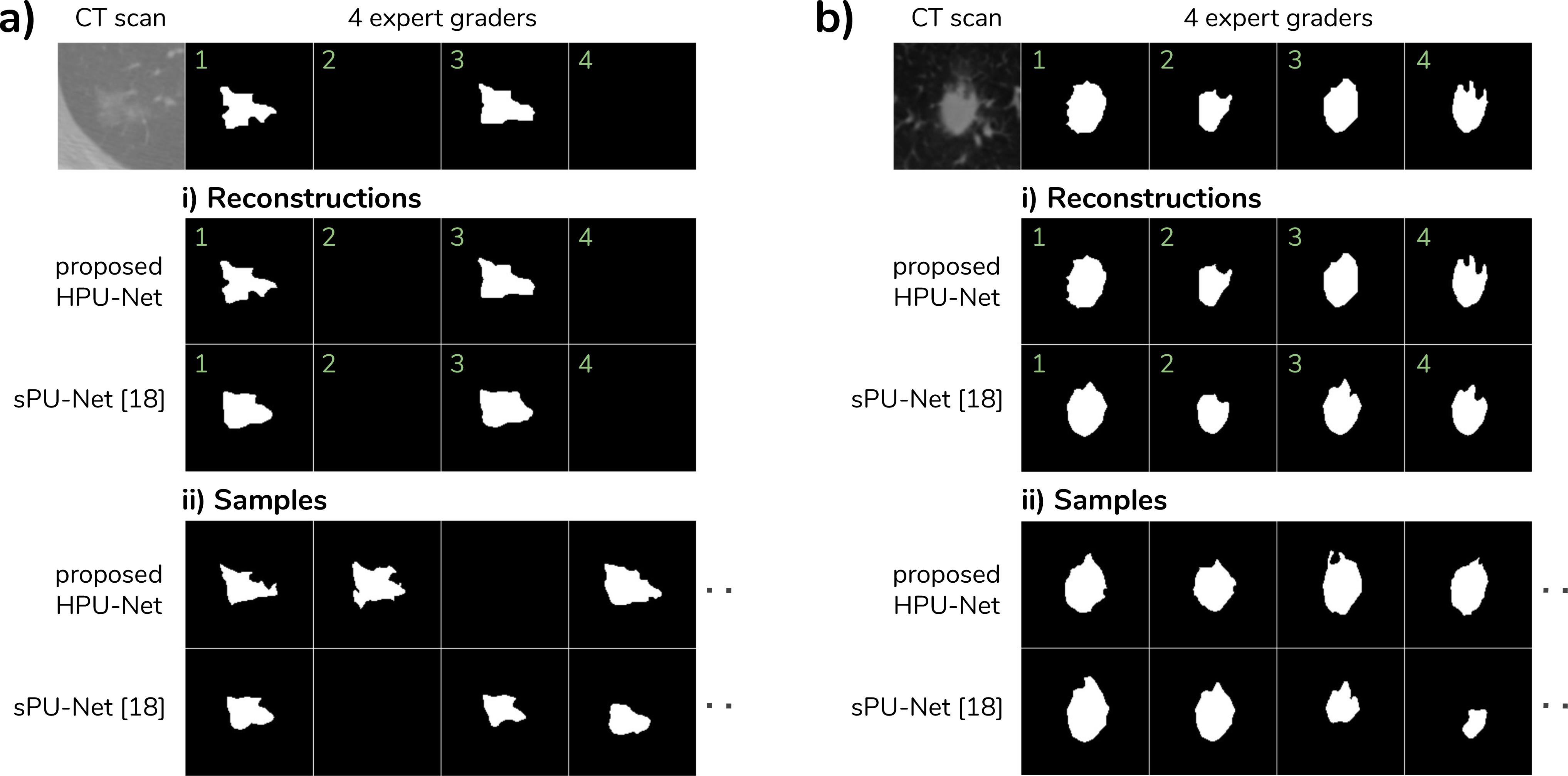}
\caption{Two example CT scans with the 4 available expert gradings from LIDC-IDRI. (\textbf{i}) Reconstructions of the 4 graders and (\textbf{ii}) Sampled segmentations. Note that the gradings can be empty, as foreground annotations correspond to supposed abnormal cases only. More cases in \autoref{appendix:hierarch_lidc_samples} and \ref{appendix:standard_lidc_samples}.}
\label{fig:lidc}
\end{figure}

The \hpu's capacity to faithfully learn segmentation distributions with high reconstruction and sample fidelity is also qualitatively evident. \autoref{fig:lidc} compares samples from both models given a pair of CT scans of prospective lung abnormalities. The hierarchical model exhibits enhanced local segmentation structure. Its samples reflect the difficulty to pin-down the boundary of normal vs. abnormal tissue from the image alone (\autoref{fig:lidc}a) and also whether or not the salient structure is abnormal. The \pu's samples on the other hand appear much coarser and `blobby' (\autoref{fig:lidc}b). 
%
\begin{figure}[tp]
\centering
\includegraphics[width=\textwidth]{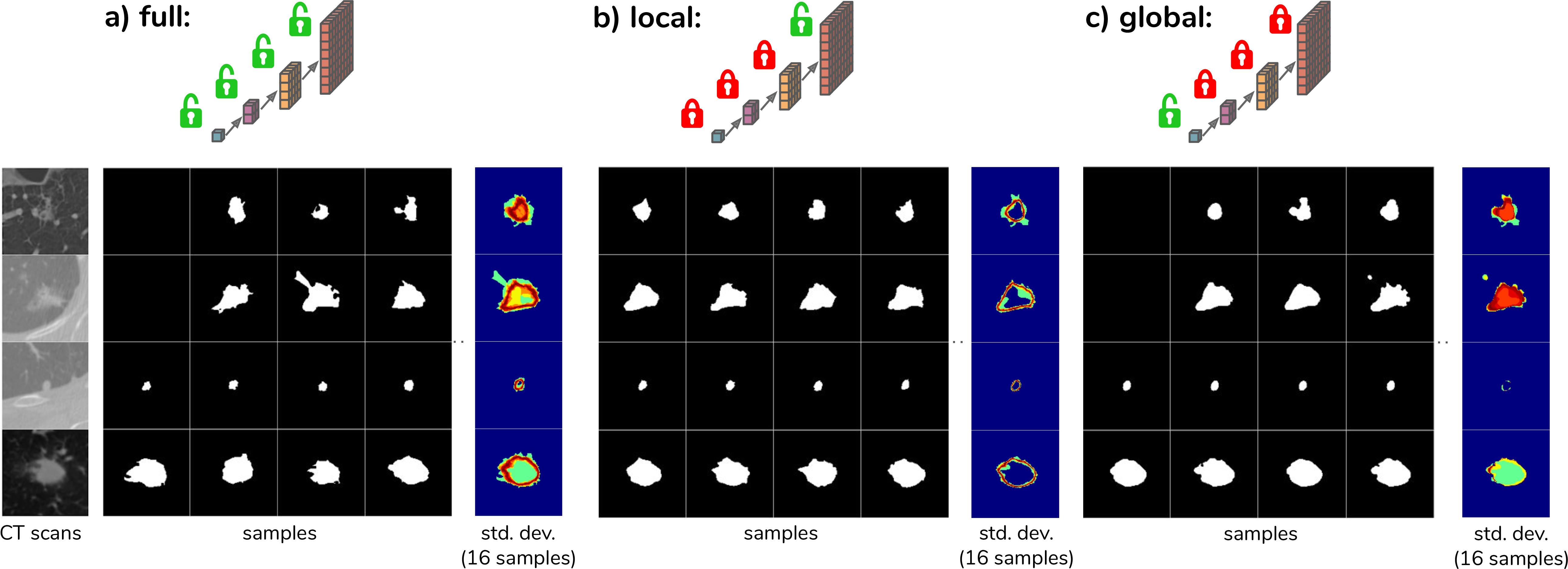}
\caption{\hpu~ samples and standard deviations across 16 samples given the CT scans on the left. Sampling from (\textbf{a}) the full hierarchy, (\textbf{b}) from only the most local latent scale  and (\textbf{c}) from only the most global scale while fixing the respectively remaining scales to their predicted means $\vec{\mu}^{\textrm{prior}}_i$. Observe in the standard deviations how the local latents alter fine details, mostly at the boundaries, while the global latents can flick the presence of coarser abnormality segmentations on and off.
}
\label{fig:lidc_scales}
\end{figure}
In order to explore how the model leverages the hierarchical latent space decomposition, we can use the predicted means $\vec{\mu}^{\textrm{prior}}_i$ for some scales instead of sampling. \autoref{fig:lidc_scales}a shows samples for the given CT scans resulting from the process of sampling from the full hierarchy, i.e. from 4 scales in this case. \autoref{fig:lidc_scales}b,c show the resulting samples when sampling from the most global or most local scale only. The hierarchical latent space appears to induce the anticipated bias: the global scales determine the coarse structure, which in this case includes the decision on whether or not the structure at hand is abnormal, while the more local scales fill in appropriate local annotations.

\subsection{SNEMI3D: Generative Instance Segmentation of Neurites}

\begin{figure}[!h]
\centering
\includegraphics[width=0.9\textwidth]{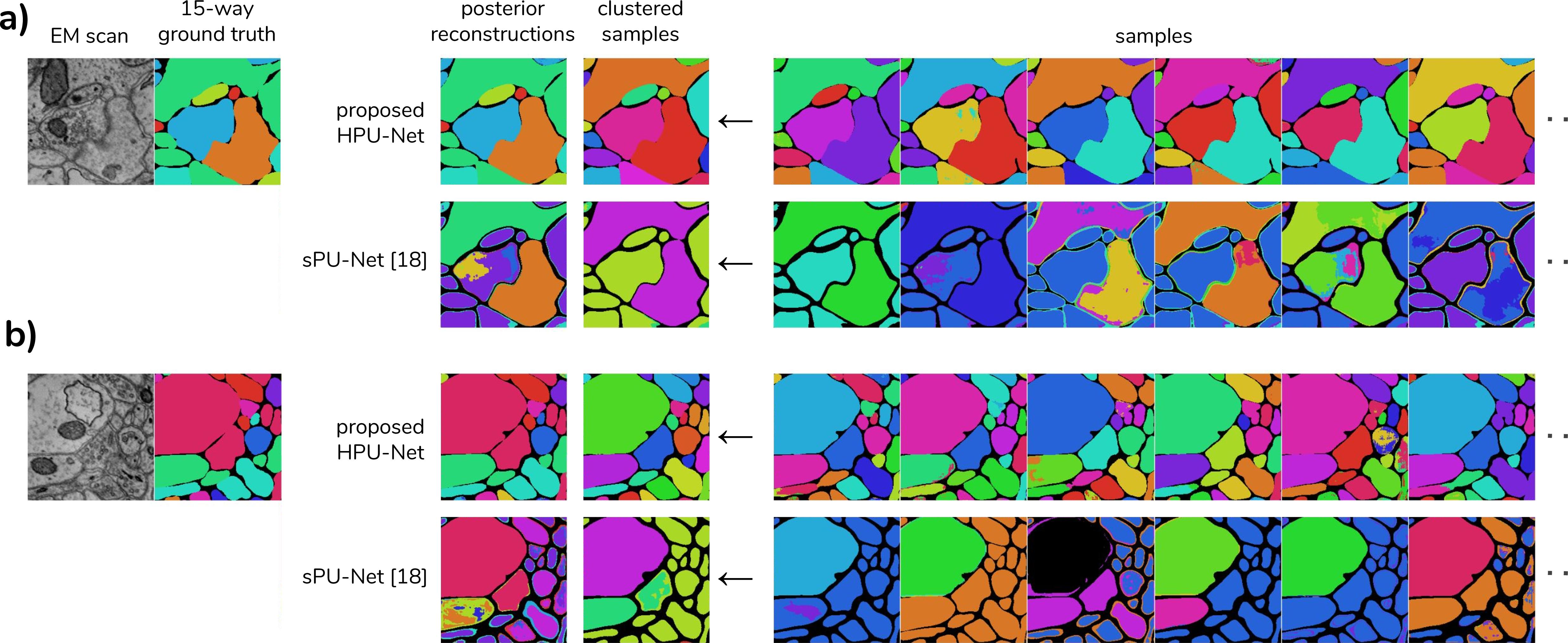}
\caption{Instance segmentation of neurons. From left to right: EM images from SNEMI3D, the ground-truth mapped to 15 random instance ids, the corresponding posterior reconstructions, predicted instance segmentation after clustering as well as 6 samples.
Color denotes instance id (one of 15) and background is shown in black. For more examples see \autoref{appendix:hierarch_snemi3d_samples} and \ref{appendix:standard_snemi3d_samples} in the appendix.
}
\label{fig:snemi3d}
\end{figure}

As a second dataset we use the SNEMI3D challenge dataset that is comprised of a fully annotated 3D block of a sub-volume of mouse neocortex, imaged slice by slice with an electron microscope \cite{kasthuri2015saturated}. We crop 2D patches of size $256 \times 256$ resulting in 1280 images for training, 160 for validation and 160 for testing (for more details see \autoref{appendix:data}). During training we randomly map the instance ids of the cells to one of 15 labels. Because the number of individual cells per image can surmount this number, the training task does not necessitate a unique predicted instance id for every cell, which is why we aggregate a number of samples for a given image to obtain a final instance segmentation. For this purpose we employ a greedy Hamming distance based clustering across 16 samples followed by a light-weight post-processing, detailed in Algorithm \autoref{algorithm:clustering} and \autoref{appendix:postprocessing}.

The \hpu, in this case using four latent scales, displays both a strong reconstruction fidelity, $\textrm{IoU}_{\textrm{rec}} = 0.60$, as well as a very low $\textrm{Rand error}=0.06$. Although we want to caution against a direct comparison between results obtained on our smaller test set (in 2D) against those from the official test set (in 3D), it is interesting to put an eye on the official leader-board, where the best dedicated algorithms reach a Rand Error of $\sim 0.025$ (e.g. \cite{lee2017superhuman}) and the human baseline achieved a value of $0.059$\footnote{\url{http://brainiac2.mit.edu/SNEMI3D/leaders-board}}. For the \pu~the parameter settings used in \cite{kohl2018probabilistic} (on other datasets) did not produce satisfactory results. Even when matching the number of global latents of a 4-scale \hpu~($\sum_{i=0}^3 2^{2i}=85$), the \pu~struggles with reconstructing instance segmentations of neurites and likewise scores badly in terms of the Rand Error, see \autoref{tab:results}c).

From \autoref{fig:snemi3d} it is evident that the \hpu~is able to sample coherent instance segmentations of these amorphous structures with largely varying size and shape, resulting in faithful instance segmentations when clustered across samples. In contrast, the \pu~has a hard time accommodating for the independently varying instances and also fails to coherently segment individual instances which is apparent in its samples, the clustering thereof and its reconstructions. 

\begin{figure}[!h]
\centering
\includegraphics[width=0.8\textwidth]{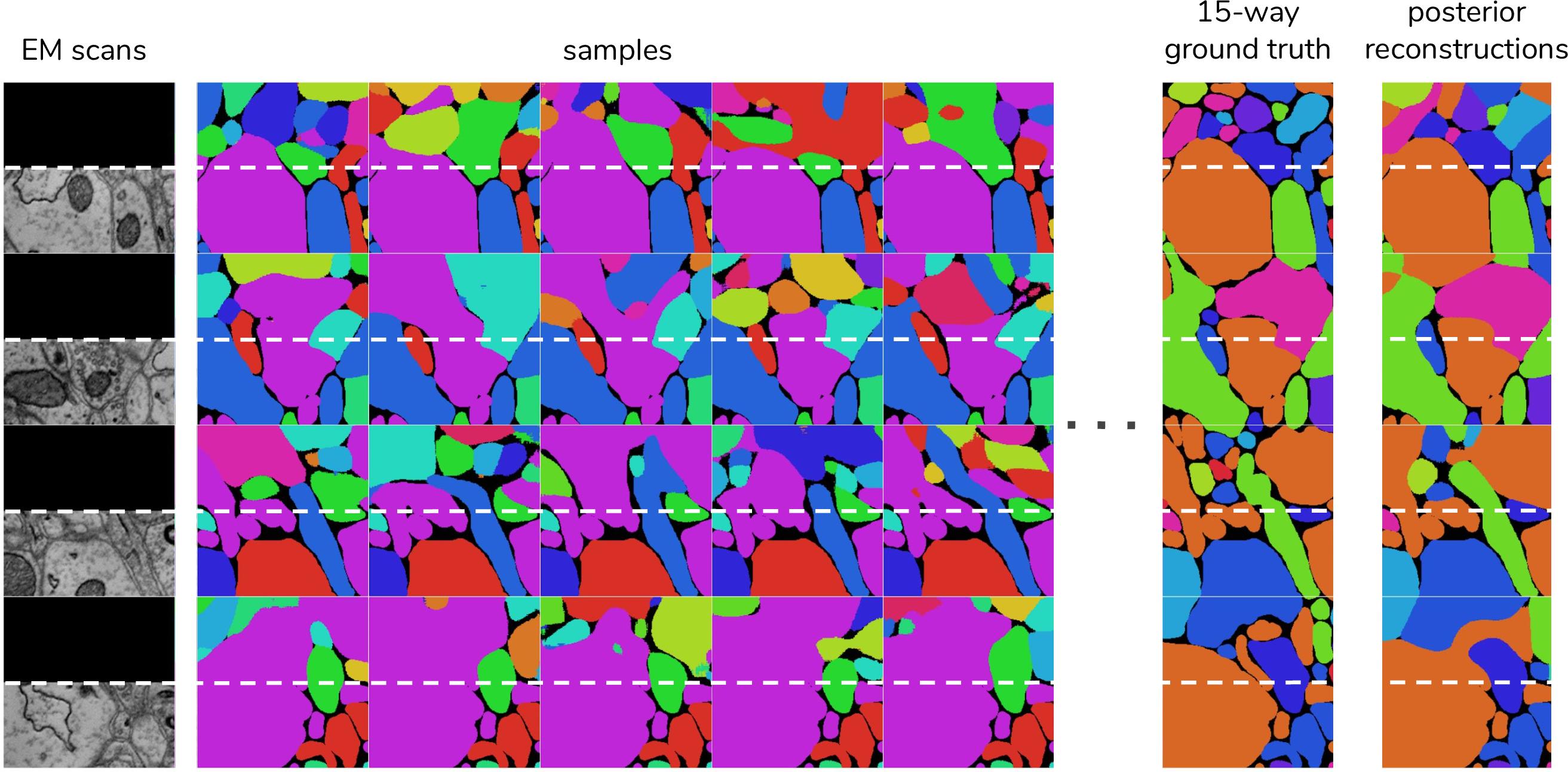}
\caption{Generative extrapolation on masked EM images with the \hpu. Areas above the dashed line in each row correspond to the masked part. Colors denote instance ids (one of 15) with black for background segmentation.}
\label{fig:snemi3d_extrapolation}
\end{figure}

\paragraph{Extrapolation Task} In order to further explore the expressiveness of the proposed generative model, we train it to generate extrapolated segmentations given masked images. The masked parts are maximally ambiguous and sensible ways of extrapolating need to be inferred from the unmasked regions. Samples and reconstructions are shown in \autoref{fig:snemi3d_extrapolation}. To be able to visualize the extrapolations across samples we feed in both the image and the ground-truth segmentation of the unmasked region to the prior, so that it can fix the found instance ids (which is not required for this to work). We observe that the model's generative structure can produce convincing extrapolations, note how the model preserves scale and appearance of unmasked instances, e.g. large cells are more likely to cover larger areas in the masked region and slim cells remain slim and elongated, see third row of samples.

\subsection{Cityscapes Cars: Generative Instance Segmentation of Cars}
\label{sec:cityscapes_results}

\begin{figure}[!h]
\centering
\includegraphics[width=\textwidth]{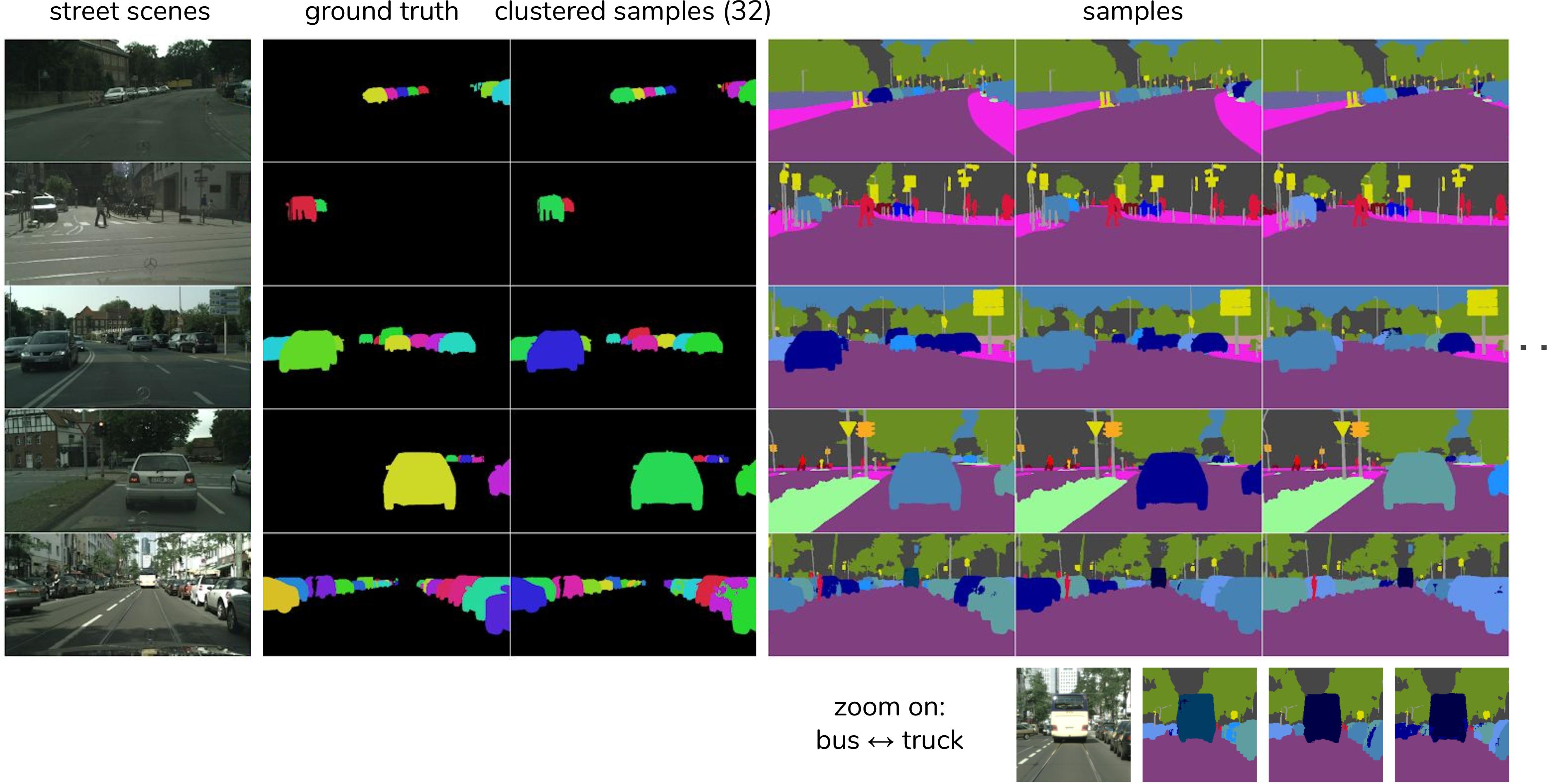}
\caption{Generative instance segmentation of cars on our Cityscapes test set on $512 \times 1024$ resolution with the \hpu. Last row that shows crops zoomed in on samples from above. More examples can be found in \autoref{appendix:hierarch_cs_cars_samples} and \ref{appendix:hierarch_cs_cars_samples_difficult}.
}
\label{fig:cityscapes_instances}
\end{figure}

In order to test our model's ability to coherently flip independent regions on natural images, we evaluate it on the task of segmenting car instances on Cityscapes. We train our model to segment all 19 Cityscapes classes while introducing additional alternative car classes that are randomly flipped during training. 
We run on half-resolution, i.e. $512 \times 1024$ (more details in \autoref{appendix:data} and \ref{appendix:training_and_architecture}). At test time we cluster 32 samples per image (see \autoref{appendix:clustering}). Results on the official validation set (our held-out test set) are reported in \autoref{tab:results}d). Employing 4 latent scales (with the highest latent resolution at $16 \times 32$), we reach an $\textrm{IoU}_{\textrm{rec}} = 0.62$, a $\textrm{Rand error} = 0.12$ and a $\textrm{AP}_{50} = 46.8$, without ensembling or any other test-time augmentations. While these results are not competitive with top-performing bounding box regressors such as Mask R-CNN \cite{he2017mask} ($\textrm{AP}_{50} = 68.3$ on the Cityscapes test set), which are tailored towards instance segmentation of boxy objects, we observe an arguably solid out-of-the-box performance of the \hpu. Direct comparison may further suffer from the post-hoc computation of object-level confidence scores which we are required to carry out for the $\textrm{AP}$-metric and which bounding-box regressors on the other hand can optimize for during training.

\autoref{fig:cityscapes_instances} shows predicted and ground truth instance segmentations for five scenes. This task is difficult as aside from varying factors such as appearance and illumination, cars nestled along the road can be heavily  occluded and individual cars can cover anything between tiny to large regions of the image. Nonetheless our proposed model can sample individual instance segmentations with good coherence, resulting in strong instance segmentations. Interestingly the model also picks up on ambiguity that is naturally present in the data, e.g. the samples in the first row show coherent flips between parts of the $road$ and $sidewalk$ and the last row shows coherent flips between $bus$ and $truck$ annotation for the bus at the end of the road (for which we provide zoomed crops of size $200 \times 200$ in \autoref{fig:cityscapes_instances}). \autoref{appendix:cs_cars_hierarchy} shows samples and the standard deviations when sampling from only the most local versus only the most global latent scales. It is apparent that the local latents affect small and distant cars while the global latents control more global factors such as cars close to the observer. This shows that also on this large scale natural image data, the model has learned to separate scales. 

\section{Discussion}
\label{sec:discussion}

Targeted at the segmentation of ambiguous medical scans, prior work  (the \pu~\cite{kohl2018probabilistic}) learns an image-global distribution that allows to sample consistent segmentation hypotheses. As we show here, this model however suffers from poor sample and reconstruction fidelity and breaks down altogether in more complex scenarios such as instance segmentation. This work proposes a much improved model, the \hpu, which shows clear quantitative and qualitative evidence for its advantages over the prior art. Our proposed model uses a much more flexible generative model and further profits from advances such as improved training procedures for VAEs and efficient hard-negative mining (which we ablate in \autoref{appendix:ablation}). The hierarchical latent space formulation enables to model ambiguities at all scales and affords the learning of complex output interdependencies such as e.g. coherent regions of pixels as found in the task of instance segmentation.

In addition to presenting high-quality results on the segmentation of ambiguous lung CT scans, we achieve strong out of the box performance in instance segmentation of both neurobiological images as well as natural images of street scenes, showing the flexibility and amenability of the proposed model to such tasks. While state-of-the-art deterministic bounding-box regressors \cite{he2017mask, lin2017focal} still perform significantly better on car instance segmentation, they are predominantly based on a pixel-wise refinement of bounding-boxes and are not designed for overlapping or intertwined instances as found in neurobiological instances. Our generative approach could be a way to directly perform dense object-level segmentation, which has recently attracted attention \cite{kirillov2018panoptic, kulikov2018instance, kulikov2019instance, xiong2019upsnet, chen2019tensormask}.

The \hpu's samples are indicative of model uncertainty for ambiguous cases that it has seen during training, which is expected to benefit prospective down-stream tasks. As such the expressed model uncertainty is valid within the data distribution only and, like many others, the model is not aware if and when it fails out-of-distribution \cite{nalisnick2018deep}. Aside from allowing to capture multiple scales of variations simultaneously, the latent hierarchy further imposes an inductive bias that mirrors the structure of many medical imaging problems, in which global information can affect top-down decision making, i.e. local annotations in our case. We show this trait in our lung CT scan experiments, where the model learns to separate variations at different scales. Here our model automatically opts to take the decision as to whether the given structure may be abnormal at its most global scale, while reserving more local decisions for local latents, see \autoref{fig:lidc_scales}. A similar decomposition is apparent on natural images (\autoref{appendix:cs_cars_hierarchy}). In terms of KL cost, it is more expensive to model global aspects locally, which in combination with the hierarchical model formulation itself, is the mechanism that puts into effect the separation of scales. Disentangled representations are regarded highly desirable across the board and the proposed model may thus also be interesting for other down-stream applications or image-to-image translation tasks.

In the medical domain the \hpu~could be applied in interactive clinical scenarios where a clinician could either pick from a set of likely segmentation hypotheses or may interact with its flexible latent space to quickly obtain the desired results. The model's ability to faithfully extrapolate conditioned on prior observations could further be employed in spatio-temporal predictions, such as e.g. predicting tumor therapy response.

\subsubsection*{Acknowledgments}

We would like to thank Peter Li and Jeremy Maitin-Shepard of Google Brain for their help with the SNEMI3D dataset, Tamas Berghammer and Clemens Meyer for engineering support and program management respectively and Jeffrey de Fauw and Shakir Mohamed for valuable discussions.

\medskip

\small
\bibliographystyle{splncs}
\bibliography{egbib}
\clearpage

\appendix
\section*{Appendices}

\section{KL-Divergence between Posterior and Prior}
\label{appendix:KL}
The Kullback-Leibler terms in \autoref{eq:elbo} and \ref{eq:geco} come about as follows:
\begin{flalign}
  D_{\mathrm{KL}}(Q||P) &= \mathbb{E}_{\vec{z} \sim Q} \left[\mathrm{log}\, Q- \mathrm{log}\, P\right] &\\
  &= \int_{\vec{z}_0,...,\vec{z}_L} \prod_{j=0}^L q(\vec{z}_j|\vec{z}_{<j})\sum_{i=0}^L \left[ \textrm{log}\; q(\vec{z}_i|\vec{z}_{<i}) - \textrm{log} \; p(\vec{z}_i|\vec{z}_{<i})\right] d\vec{z}_0 ... d\vec{z}_L, \\
   &\text{using} \int \phi(\vec{z}_i)\; \prod_{j=0}^L q(\vec{z}_j|\vec{z}_{<j}) d\vec{z}_0 ... d\vec{z}_L = \int  \phi(\vec{z}_i)\; \prod_{j=0}^i q(\vec{z}_j|\vec{z}_{<j}) d\vec{z}_0 ... d\vec{z}_i:\\
  &= \sum_{i=0}^L \int_{\vec{z}_0, ..., \vec{z}_i} \prod_{j=0}^{i} q(\vec{z}_j|\vec{z}_{<j}) \left[ \textrm{log}\; q(\vec{z}_i|\vec{z}_{<i}) - \textrm{log} \; p(\vec{z}_i|\vec{z}_{<i})\right] d\vec{z}_0 ... d\vec{z}_i \\
  &= \sum_{i=0}^L \int_{\vec{z}_0, ..., \vec{z}_i} \prod_{j=0}^{i-1} q(\vec{z}_j|\vec{z}_{<j})\; q(\vec{z}_i|\vec{z}_{<i}) \left[ \textrm{log}\; q(\vec{z}_i|\vec{z}_{<i}) - \textrm{log} \; p(\vec{z}_i|\vec{z}_{<i})\right] d\vec{z}_0 ... d\vec{z}_i \\
  &= \sum_{i=0}^L \mathbb{E}_{\vec{z}_{<i} \sim Q} D_{\mathrm{KL}}(q(\vec{z}_i|\vec{z}_{<i}) || p(\vec{z}_i|\vec{z}_{<i})),
\end{flalign}
where for improved clarity we omit $X$ and $X, Y$ as conditional arguments to $p$ and $q$ respectively. For brevity our notation additionally subsumes $q(\vec{z}_0) := q(\vec{z}_0|\vec{z}_{-1})$ and similar for $p(\vec{z}_0)$. For our choice of posterior and prior distribution (see \autoref{eq:prior} and \ref{eq:posterior}) the KL-terms above can be evaluated analytically. The expectations in \autoref{eq:elbo} and \ref{eq:geco} using samples $\vec{z} \sim Q$ are performed with a single sampling pass.

\section{Performance Measures}
\label{appendix:metrics}

\subsection{Distribution Agreement}

We report how well the distribution produced by the respective generative model and the given ground-truth distribution agree on the LIDC dataset. In real world scenarios such as the LIDC dataset, the ground-truth distribution is only known in terms of a set of samples. One way to measure the agreement between two distributions that only requires samples as opposed to knowledge of the full distributions is by means of the \textbf{Generalized Energy Distance} ($\textrm{GED}^2$, also referred to as Maximum Mean Discrepency). This kernel-based metric was employed in \cite{kohl2018probabilistic} using $1 - \textrm{IoU}(Y, Y')$ as a distance kernel, where IoU is the intersection over union metric between two segmentations. We found this measure inadequate in such cases where the models' samples only poorly match the ground truth samples, since the metric then unduly rewards sample diversity, regardless of the samples' adequacy. As an alternative that appears less vulnerable to such pathological cases, we propose to use the Hungarian algorithm \cite{kuhn1955hungarian, munkres1957algorithms} to match samples of the model and the ground-truth. The Hungarian algorithm finds the optimal 1:1-matching between the objects of two sets, for which we use $\textrm{IoU}(Y, Y')$ to determine the similarity of two samples. We report the match as the \textbf{Hungarian-matched IoU}, i.e. the average IoU of all matched pairs and duplicate both sets so that their number of elements matches their least common multiple. As empty segmentations can be valid gradings in the LIDC dataset we need to define how the IoU enters the distribution metrics for the case of correctly predicted absences, which is detailed below.

\subsection{Reconstruction Fidelity}

The reconstruction fidelity is an upper bound to the fidelity of the conditional samples. In order to asses this upper bound on the fidelity of the produced segmentations we measure how well the model's posteriors are able to reconstruct a given segmentation in terms of the IoU metric, i.e. we report the \textbf{reconstruction IoU},  $\textrm{IoU}_{\textrm{rec}}(Y, Y')$ where $Y' = S(X, \vec{\mu}^{\textrm{post}}(X,Y))$. Whenever we employ the IoU-metric, i.e. also when it enters the measures for distribution agreement, we calculate it with respect to the stochastic foreground classes only. We further do not calculate it globally across all the test set pixels (as is regularly done in semantic segmentation challenges, e.g. in Cityscapes \cite{cordts2016cityscapes}), but calculate it across the pixels of each image and then average across all test set images. For these reasons the question arises how to deal with a correctly predicted absence of a class in an image, a case for which the IoU metric is undefined (the denominator would be 0). For the LIDC dataset, empty ground-truth segmentations can be a valid grading which is why we follow \cite{kohl2018probabilistic} and define a correctly predicted absence as IoU = 1. In the SNEMI3D and Cityscapes instance segmentation tasks we do not want to evaluate whether a model correctly predicts a class' absence, which is why we correct class absences do not enter the mean IoU of an image, while wrongly predicted absences are penalized (in practice we perform a `NAN-mean' over the classes of interest). In the Cityscapes case we additionally make use of the provided ignore-masks, keeping unlabeled pixels out of the evaluations.

\subsection{Instance Segmentation}

In order to score how well the predicted instance segmentations (the instance clusters) agree with the ground truth, we calculate the \textbf{Rand Error}. This measure is defined as $1 -F$-score, where the precision and recall values that enter the $F$-score are determined from whether pixel pairs between the ground truth clustering and a predicted clustering belong to the same segment (positive class) or different segments (negative class) \cite{rand1971objective, arganda2015crowdsourcing}. We use the foreground-restricted version as employed in the SNEMI3D challenge\footnote{Available as {\tt{adapted\_rand\_error}} in the python package \href{https://gala.readthedocs.io/en/latest/api/gala.evaluate.html}{gala} \cite{nunez2014graph}.}.

On Cityscapes instance segmentation we additionally report the \textbf{Average Precision} (AP). It is based on object level scoring and defined as the area under the precision recall curve for all predicted object detections. To span the precison recall curve, an object level score that quantifies a model's confidence in the `objectness' of its prediction is required. For our car instance segmentation experiments we employ the Cityscapes evaluation scheme\footnote{Official evaluation code can be found \href{https://github.com/mcordts/cityscapesScripts}{here}.}, reporting $\textrm{AP}_{50}$ and $\textrm{AP}$, the average precision when requiring predictions to match above a thresholded $\textrm{IoU}_{\textrm{thres}} > 0.50$ and when averaging across multiple such thresholds (10 different overlaps ranging from 0.5 to 0.95 in steps of 0.05), respectively. To artificially obtain object-level scores we average the softmax scores of all stochastic classes across samples and pixels of a predicted instance mask \cite{kulikov2018instance}.

\section{Dataset Details}
\label{appendix:data}

\textbf{LIDC-IDRI} The LIDC-IDRI dataset \cite{armato2015, armato2011lung, clark2013cancer} contains 1018 lung CT scans from 1010 lung patients with manual lesion segmentations from four experts. We use the same setup as in \cite{kohl2018probabilistic}, i.e. we employ the annotations from a second reading and employ the same data splits (722 patients for training, 144 patients for validation and 144 patients for the test set). The data is then cropped to 2D images of shape $180 \times 180$ pixels, resulting in 8882 images in the training set, 1996 images in the validation set and 1992 images in the test set. Crops are centered at positions for which at least one grader indicates a lesion. In the LIDC dataset, empty foreground segmentations can be viable expert gradings, which is why we employ $\textrm{IoU} = 1$ when a model sample agrees in such cases. $\textrm{IoU}_{\textrm{rec}}$ is an average of the reconstructions of all four gradings. As in \cite{kohl2018probabilistic} we only use those lesions that were specified as a polygon (outline) in the XML files of the LIDC dataset, disregarding the ones that only have center of shape. That is, according to the LIDC paper only such lesions that are larger than 3mm are used, with smaller ones filtered out as they are regarded clinically less relevant \cite{armato2011lung}. We filter out each Dicom file whose absolute value of SliceLocation differs from the absolute value of ImagePositionPatient[-1]. Finally we assume that two masks from different graders correspond to the same lesion if their tightest bounding boxes overlap.

\textbf{LIDC-IDRI subset B} For `Subset B' we consider only those test set cases, which have annotations by all 4 graders, i.e. all graders agree on the presence of an abnormality. This results in 638 images, so close to a third of the full test set.

\textbf{SNEMI3D} As a second dataset we use the SNEMI3D challenge\footnote{\url{http://brainiac2.mit.edu/SNEMI3D/home}} dataset that is comprised of a fully annotated 3D block of a sub-volume of mouse neocortex, imaged slice by slice with an electron microscope \cite{kasthuri2015saturated}. This stack is $1024 \times 1024 \times 100$ voxels large, comes at a voxel size of $6 \times 6 \times \SI{29}{\nano\meter}^3$ and contains a total of 400 fully annotated neurite instance annotations. We use the first 80 z-slices as our training dataset, the adjacent 10 slices as a validation set and the remaining 10 slices as a test set to report results on. We crop non-overlapping patches of size $256 \times 256 \times 1$ resulting in 1280 images for training, 160 for validation and 160 for testing. During training we randomly map the instance identifiers (ids) of the cells to one of 15 labels, thereby treating the instance id of the cells as latent information that the networks need to model. Because the number of individual cells per image can surmount this number, the training task does not necessitate a unique predicted instance id for every cell. This means that in order to obtain a predicted instance segmentation at test time, we need to aggregate a number of samples for a given image. For this purpose we employ a greedy Hamming distance based clustering across $n$ samples followed by a light-weight post-processing, detailed in Algorithm \ref{algorithm:clustering} of \autoref{appendix:clustering} and \autoref{appendix:postprocessing} (we chose $n=16$ and Hamming distance threshold $\alpha=16$).

\textbf{Cityscapes} As a third dataset we use the Cityscapes street scene dataset that comes with both dense category segmentations, as well as with instance segmentations for a number of categories. The official Cityscapes training dataset (with fine annotations) comprises 2975 images. We employ the official validation set of 500 images as a test set to report results on, and split off 274 images (corresponding to the 3 cities of Darmstadt, Mönchengladbach and Ulm) from the official training set as an internal validation set. At test time, we cluster 32 samples per image (see \autoref{appendix:clustering}), using a threshold of $\alpha=32$.

\section{Architecture and Training Details}
\label{appendix:training_and_architecture}

\textbf{Architecture} The Hierarchical Probabilistic U-Net implementations employed for the different tasks differ in their number of processing scales, the depth of each such scale and the number of latent scales employed. The architecture as employed during the sampling process is depicted in \autoref{fig:architecture}a. The setup for the training process is depicted in \autoref{fig:architecture}b) and includes the separate posterior net that is required during training.

At each processing scale of the \hpu~we employ a stack of $n$ pre-activated residual blocks \cite{he2016identity} (grey blocks in \autoref{fig:architecture}), where $n$ determines the depth of that scale. For both the LIDC and SNEMI3D experiments we use $n=3$ residual blocks and for the Cityscapes experiment we use $n=2$ residual blocks at each processing scale of the U-Net's encoder and decoder respectively. Similar to \cite{kingma1606improving, maaloe2019biva}, we find the use of unobstructed connections (in our case res-blocks) between latent scales of the hierarchy to be crucial for the lower scales to be employed by the generative model. Without the use of res-blocks the KL-terms between distributions (indicated by green connecting lines in \autoref{fig:architecture}) at the beginning of the hierarchy often become $\sim0$ early on in the training, essentially resulting in uninformative and thus unused latents.

In each res-block the residual feature map is calculated by means of a series of three $3 \times 3$-convolutions, the first of which always halves the number of the feature maps employed at the present scale, such that the residual representations live on a lower dimensional manifold. At the end of the residual branch a single (un-activated) $1 \times 1$-convolution projects the features back to the number of features of the given scale. The resulting residual is then added to the skipped feature map, which is skipped forward (i.e. left untouched) unless the number of feature maps is set to change, in which case it is projected by a $1 \times 1$-convolution. This happens only at transitions that change the feature map resolutions. For down-sampling of feature maps we use average pooling and upsample by using nearest neighbour interpolation. As described in \autoref{sec:architecture}, the spatial grid of latent variables is sampled at the end of each U-Net decoder scale that is part of the hierarchy and concatenated to the final feature map produced at this scale, before both are up-sampled.

The number of latent scales is chosen empirically such as to allow for a sufficiently granular effect of the latent hierarchy. For the tasks and image resolutions considered here, we found 3 - 5 latent scales to work well. The number of processing scales is chosen such that a smallest possible spatial resolution is achieved in the bottom of the U-Net. For the square images in LIDC and SNEMI3D this means a resolution of $1 \times 1$ and for the Cityscapes task the minimum resolution is $1 \times 2$ (in this case we however employ $2 \times 4$, which is detailed below). The employed separate posterior mirrors the number of scales and the number of feature maps of the corresponding components in the U-Net, see the bottom part of \autoref{fig:architecture}b. Its only architectural difference is its first convolutional layer, which processes the input image concatenated with the corresponding one-hot segmentation along the channel axis. All weights of all models are initialized with orthogonal initialization having the gain (multiplicative factor) set to 1, and the bias terms are initialized by sampling from a truncated normal with $\sigma = 0.001$.

\textbf{Training} The \hpu~is trained using the GECO-objective (\autoref{eq:geco}) and a stochastic top-k reconstruction loss. As described in \autoref{sec:architecture}, the $k$th percentile employed for the top-k objective is fixed across tasks to 2$\%$ of each batch's pixels. The GECO-objective aims at matching a reconstruction target value $\kappa$. For each experiment we chose $\kappa$ sufficiently low so as to correspond to a strong reconstruction performance while resulting in a training schedule that is not dominated by the reconstruction term over the entire course of the training (e.g. if $\kappa$ is chosen too high, the Lagrange multiplier $\lambda$, and thus the learning pressure it exerts, mounts and remains on the reconstruction term rather then moving over on the KL terms). The desired behavior of the reconstruction objective ${\cal{L}}_{\textrm{rec}}$ and the Lagrange multiplier $\lambda$ can be observed in~\autoref{fig:lidc_training} and \autoref{fig:snemi3d_training}, where $\lambda$ rises until ${\cal L}_{\textrm{rec}}$ matches $\kappa$, after which $\lambda$ drops and the pressure on the KL-terms increases.

In contrast to the regular cross-entropy employed in semantic segmentation, the reconstruction error here is not averaged but summed over individual pixels (before being averaged across batch instances). This is because the likelihood is assumed to factorize over the pixels of an image and so their log-likelihood is summed over. For comparability we however report ${\cal{L}}_{\textrm{rec}}$ and $\kappa$ per pixel (e.g. in \autoref{fig:lidc_training}, \autoref{fig:snemi3d_training} and in Table~\ref{tab:ablation}).

The precise training setups for each of the tasks and models are reported below. Note that the training objectives for all models encompass an additional weight-decay term that is weighted by a factor of $1e^{-5}$.

\subsection{LIDC-IDRI Lung CT scans}

During training on LIDC, image-grader pairs are drawn randomly. Similar to what was done in \cite{kohl2018probabilistic}, we apply random augmentations\footnote{We use the code available at \url{https://github.com/deepmind/multidim-image-augmentation/}.} to the image and label tiles ($180 \times 180$ pixels size) including random elastic deformation, rotation, mirroring, shearing, scaling and a randomly translated crop that results in a tile size of $128 \times 128$ pixels. Any padding added to the images and labels during the augmentation process is masked from the loss during training.

In order to evaluate the Probabilistic U-Net on additional metrics than those employed in \cite{kohl2018probabilistic}, we retrain a re-implementation of the model with the exact same hyperparameters and setup as in \cite{kohl2018probabilistic}, i.e we employ a 5-scale model, with three $3 \times 3$-convolutions per encoder and decoder-scale, a separate prior and posterior net that mirror the used U-Net's encoder as well as 6 global latents and three final $1 \times 1$ convolutions. Moreover we employ an identical ELBO-formulation ($\beta=1$), train with identical batch-size of 32, number of iterations ($240k$) and learning rate schedule $0.5e^{-5} \rightarrow 1e^{-6}$.

On LIDC, the \hpu~uses 8 latent scales resulting in a global $1 \times 1$-`U-Net bottom' and 3 res-blocks per encoder and decoder scale. The base number of channels is 24 and until the fourth down-sampling the number of channels is doubled after each down-sampling operation, resulting in a maximum width of 192 channels. The U-Net's decoder mirrors this setup. We train the \hpu~ with an initial learning rate of $1e^{-4}$ that is lowered to $0.5e^{-5}$ in 4 steps over the course of $240k$ iterations. The employed batch-size is 32. The \hpu~is trained with the GECO-objective using $\kappa=0.05$.

\begin{figure}[!h]
\centering
\includegraphics[width=1.0\textwidth]{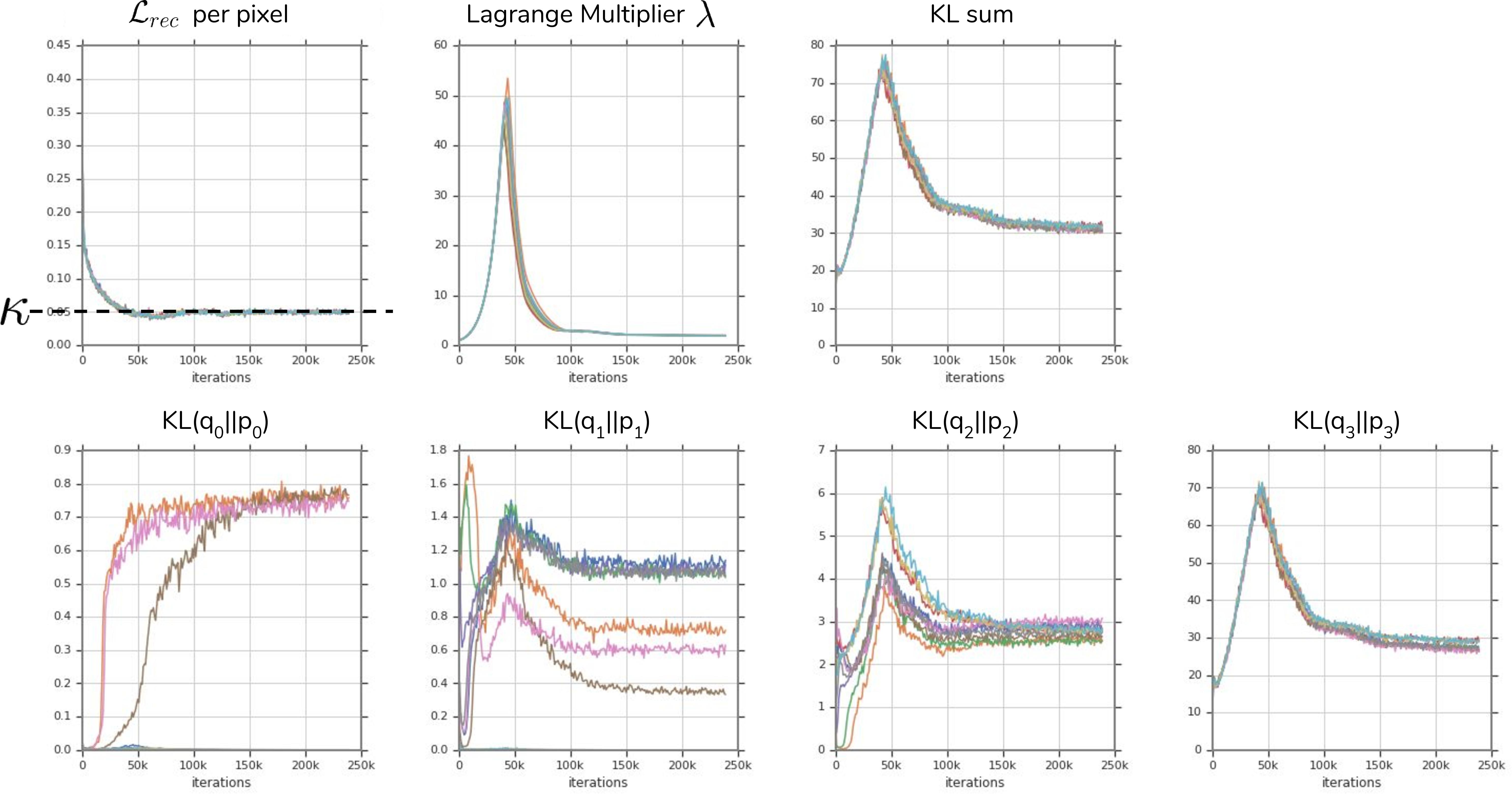}
\caption{Components of the learning objective in the course of the LIDC training for 10 random initializations.}
\label{fig:lidc_training}
\end{figure}

\autoref{fig:lidc_training} shows how the top-k reconstruction term ${\cal L}_{\textrm{rec}}$, the Lagrange multiplier $\lambda$, as well as the individual KL-terms (and their sum) progress in the course of training for the 10 random model initializations reported in \autoref{tab:results}. As mentioned above, GECO structures the dynamics such that $\lambda$ puts pressure on ${\cal L}_{\textrm{rec}}$ until it reaches its target value $\kappa$. After that the training objective holds the reconstruction term at $\kappa$ while optimizing for lower overall Kullback-Leibler divergence (`KL'). The KL is a measure for how much more information the posterior distribution carries compared to the prior, a quantity that we aim to minimize. Note that the KL-sum is very similar for all models, but the way the KL splits across the hierarchy can differ. The models that end up using the global latents profit from a slightly lower overall KL indicating that this decomposition is more efficient, e.g. it is more efficient not to repeat global information in the local latents when it is already provided by global latents etc.

\subsection{SNEMI3D neocortex EM slices}

During training on SNEMI3D we randomly sample a latent (class) id for each cell in each image. We limit the number of instance ids to 15 and just like on LIDC we apply random augmentations including random elastic deformation, rotation, mirroring, shearing, scaling and a randomly translated crop. Any padding added to the images and labels during the augmentation process is masked from the loss during training.

For the standard Probabilistic U-Net we employ a 9 scale architecture and a base number of 24 channels, that until the 4th down-sampling, is doubled after each down-sampling operation, resulting in a maximum width of 192 channels. The \pu~again uses three $3 \times 3$-convolutions per encoder and decoder scale, while the \hpu~employs three res-blocks. The \hpu~also employs 32 base channels, a total of 9 scales interleaved with four (scalar) latent scales, resulting in a total of 85 latents. This is also the number of global latents that we used for the \pu, since employing low numbers of latents, such as $\sim 10$ as proposed in \cite{kohl2018probabilistic} never converged (even working with 85 global latents does not make for a very stable training). Both models are trained for $450k$ iterations with a batch-size of 24, and an initial learning rate of $1e^{-4}$ that is lowered to $1e^{-7}$ in 5 steps. The \hpu~is trained with the GECO-objective using $\kappa=1.20$.

\begin{figure}[h]
\centering
\includegraphics[width=1.0\textwidth]{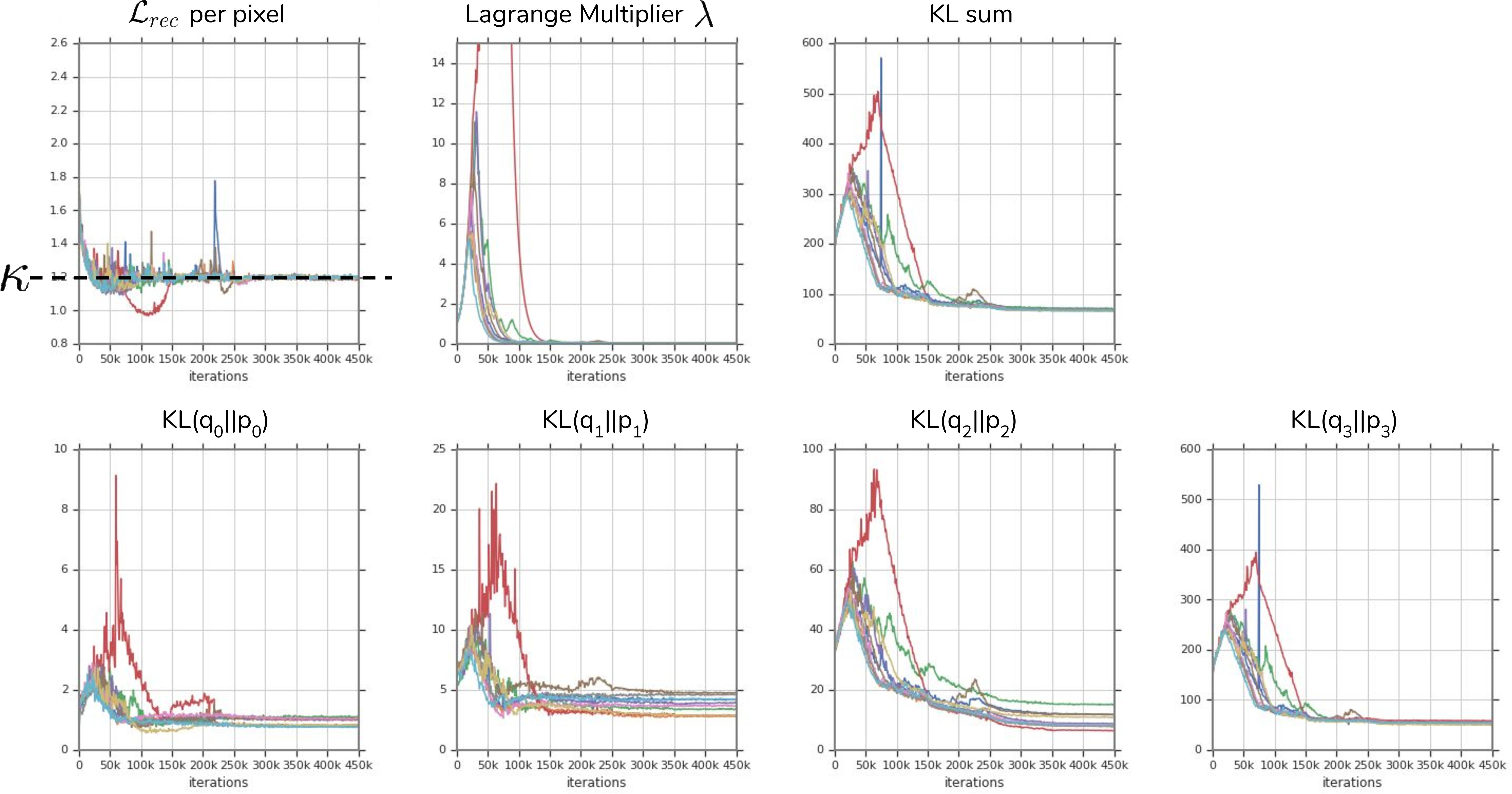}
\caption{Components of the learning objective in the course of the SNEMI3D training for 10 random initializations.}
\label{fig:snemi3d_training}
\end{figure}

\autoref{fig:snemi3d_training} again shows how the top-k reconstruction term ${\cal L}_{rec}$, the Lagrange multiplier $\lambda$, as well as the individual KL-terms (and their sum) progress in the course of training for the 10 random model initializations reported in \autoref{tab:results}. Again the KL sums to a similarly low value across models with different decompositions across the four scales.

\subsection{Cityscapes Car Instances}

We resample the Cityscapes images and labels to half-resolution, i.e. $512 \times 1024$. During training we randomly sample a (latent) instance id for each car in the image, where we limit the total number of car ids to 5. We apply random deformations including random color augmentations, elastic deformation, rotation, mirroring, shearing, scaling and a randomly translated crop. Any padding added to the images and labels during the augmentation process is masked from the loss during training alongside any such pixels that are marked as part of the `ignore'-class in the dataset (pixels that can't be attributed to one of the provided 19 classes).

We train a \hpu~with 9 scales, resulting in a $2 \times 4$-`U-Net bottom' and 4 latent scales. Using another scale (so 5 latent scales and a number of 10 overall scales) did not significantly change the results and due to the image aspect ratio of 1:2, does not result in a fully global latent scale either. The employed model uses two res-blocks for each encoder and decoder scale and we train the model with a batch-size of 128 for $100k$ iterations using TPU accelerators and spatial batch partitioning. We use an initial learning rate of $2e^{-4}$ that is halved after $70k$ iterations. The base number of channels is 32 and until the fourth down-sampling the number of channels are doubled after each down-sampling operation, resulting in a maximum width of 256 channels. The \hpu~is trained with the GECO-objective using $\kappa=0.77$.

\section{$\boldsymbol{\textrm{GED}^2}$ on LIDC subset B}
\label{appendix:ged_subset_b}
On `Subset B' the \pu~gets a $\textrm{GED}^2 = 0.52 \pm~0.09$ while the \hpu~achieves as $\mathrm{GED}^2 = 0.38 \pm~0.02$. Both values result from the set of 10 models used for the LIDC results in \autoref{tab:results} (again using 1000 bootstraps with replacement).

\section{Ablation Study}
\label{appendix:ablation}

In order to show the effect of some of the main choices we made for the model and the loss formulation, we perform an ablation study on the LIDC lung abnormalities segmentation task. All models are trained with the same training setup and hyper parameters as used in the LIDC experiments (described in \autoref{appendix:training_and_architecture}), if not stated differently in the following. First we evaluate the importance of the latent hierarchy. We train 10 random initializations for a model with a global latent scale in the `U-Net's bottom' that otherwise employs the same model topology as the \hpu~that we employ on LIDC. For this model we use 85 global latents, i.e. the same number of total latents that the 4-scale hierarchical model employs. In order to arrive at a comparable reconstruction IoU, we found it necessary to raise the reconstruction target $\kappa$ above the value of 0.05 (employed for the other models) to a value of $\kappa=0.15$. As reported in \autoref{tab:ablation}, this model performs significantly worse than the \hpu~in terms of both $\textrm{GED}^2$ and the Hungarian-matched IoU, while also suffering from a loss in reconstruction fidelity. As a second model configuration we consider a model with the same topology as the employed \hpu, however employing only its most local scale of latents (a spatial grid of size $8 \times 8$). The idea is to assess to what degree the latents lower in the hierarchy help coordinate the sampling from the last, most finely resolved grid of latents. The results in \autoref{tab:ablation} show another significant decrease in the model's ability to match the ground truth distribution, suggesting that the hierarchy indeed is an important model choice enabling the strong performance in terms of $\textrm{GED}^2$ and the Hungarian-matched IoU. Lastly we quantify the effect of employing a top-k loss for the hierarchical model. The last row in \autoref{tab:ablation} shows the positive effect that the top-k loss formulation has on the reconstruction IoU ($\textrm{IoU}_{rec}$), while allowing to keep the same level of distribution match (there is a slight increase in Hungarian-matched IoU when ablating the top-k loss, it is however insignificant across 10 random initializations).

\begin{table}[h!]
  \caption{Ablation study on LIDC-IDRI. All results are reported on our test set and the given means and standard deviations are taken across 10 random initializations of the same respective model setup and 1000 bootstraps with replacement each. The values reported for $\kappa$ are normalized per pixel and for comparison the LIDC results reported in \autoref{tab:results} are shown in the first row of this table.}
  \label{tab:ablation}
  \centering
  \scalebox{0.85}{
  \begin{tabular}{lccc}
    \toprule
    \textbf{model + loss formulation} & $\textbf{IoU}_{\textbf{rec}}$ & $\textbf{GED}^{\textbf{2}}$ & \textbf{Hungarian-matched IoU} \\
    \midrule
    \rule{0pt}{2.5ex}
    4-scale hierarchy + GECO ($\kappa$ = 0.05) + top-k (k=0.02) & 0.97 $\pm$ 0.00 & 0.27 $\pm$~0.01 & 0.53 $\pm$~0.01 \\ 
    \midrule
    \rule{0pt}{2.5ex}
    local latents + GECO ($\kappa$ = 0.05) + top-k (k=0.02) & 0.97 $\pm$ 0.00 & 0.34 $\pm$~0.01 & 0.45 $\pm$~0.01 \\ 
    \rule{0pt}{2.5ex}
    global latents + GECO ($\kappa$ = 0.15) + top-k (k=0.02) & 0.94 $\pm$~0.02 & 0.40 $\pm$~0.02 & 0.37 $\pm$~0.02\\ 
    \rule{0pt}{2.5ex}
    4-scale hierarchy + GECO ($\kappa$ = 0.05) & 0.94 $\pm$~0.00 & 0.27 $\pm$~0.01 & 0.54 $\pm$~0.01  \\ 
    \bottomrule
  \end{tabular}}
\end{table}

\section{Hamming Distance based Greedy Clustering}
\label{appendix:clustering}

\normalem

\begin{algorithm}[H]
\label{algorithm:clustering}
 \caption{Hamming distance based greedy clustering, which makes use of the assumption that pixels of the same object vary together across samples. Pseudo-code used to get instance segmentations from segmentation samples $\vec{Y}_i$ for a given image $X$ of size $H \times W$. The employed algorithm assigns pixels to clusters based on the Hamming distance between a cluster's prototype and the pixel's vector representation. The Hamming distance is a simple count of element-wise mismatches. Both vectors consist of the respective pixels' sampled class labels in one-hot form, i.e. for $n$ samples and $C$ classes, they have length $nC$. The algorithm proceeds in a greedy manner, i.e. once no more matches satisfying an upper bound on the distance to the current prototype are found, a new prototype is randomly picked from the remaining unassigned pixels. Sampling at random rather than picking the next available pixel minimizes the clustering run-time (which is ${\cal O} \leq (HW)^2$) and the likelihood of picking cluster prototypes from object boundaries. The algorithm starts with assigning pixels to a provided background class label. This assures that cluster $c = 0$ always corresponds to the background class, but is not strictly necessary, alternatively the algorithm can omit the case distinction in line 10ff.}

\SetKwInput{KwData}{Parameters}
\KwResult{Instance Segmentation $\vec{I} \in \Z^{H \times W}$.}
\KwData{$n$: number of samples, $\alpha$: threshold.}
 Retrieve $n$ sample segmentations $\vec{Y}_i^{\textrm{prob}} \in \R^{H \times W \times C};\;\; \vec{Y}_i^{\textrm{prob}}  \leftarrow S(X, \vec{z}_i) , \;\vec{z}_i \sim P(.|X)$;\\
 Transform samples to one-hot $\vec{Y}_i \in \Z^{H \times W \times C} $, $\vec{Y}_i \leftarrow \textrm{one\_hot}(\textrm{argmax}(\vec{Y_i}^{\textrm{prob}})) $;\\
 Concatenate samples over channels $\vec{Y} \in \Z^{H \times W \times nC};\;\; \vec{Y} \leftarrow \textrm{concat}([\vec{Y}_0,...,\vec{Y}_n]) $;\\
 Initialize Instance Segmentation $\vec{I} \in \Z^{H \times W};\;\; \vec{I} \leftarrow [[-1,\dots],\dots] $;\\
 Inititalize set of unassigned pixels $\mathcal{U} = \textrm{where}(\vec{I} == -1)$;\\
 Initialize background one-hot vector $\vec{b} \in \Z^{C \times 1};\;\; \vec{b} \leftarrow \textrm{one\_hot}(\textrm{background label}) $;\\
 Initialize protoype $\vec{p} \in \Z^{nC \times 1}$ with the prototype of the background class $\vec{p}  \leftarrow \textrm{concat}([\vec{b}, \vec{b}, \dots]) $;\\
 Inititialize cluster id $c = 0$;\\
 \While{$|\mathcal{U}| > 0$}{

  \eIf{c==0}{
   Do nothing, as $\vec{p}$ is initially assigned to background class prototype\;
  }{
  Draw a random pixel from the set of unassigned pixels $i \sim \mathcal{U}$\;
  Use the one-hot sample vector of this pixel as the $c$th cluster's prototype $\vec{p} \leftarrow \vec{Y}[i]$\;
  $\vec{I}[i] \leftarrow c$\;
  Drop $i$ from set of unassigned pixels $\mathcal{U} \leftarrow \mathcal{U} \setminus \{i\}$\;
  }
  \ForEach{ $j \in \mathcal{U}$}{
  Retrieve one-hot sample vector of the pixel $j$ as $\vec{\nu} \leftarrow \vec{Y}[j]$\;
  Calculate Hamming distance $d = \textrm{hamming\_distance}(\vec{\nu}, \vec{p})$\;
  \If{$d \leq \alpha$}{
  $\vec{I}[j] \leftarrow c$\;
  $\mathcal{U} \leftarrow \mathcal{U} \setminus \{i\}$\;}
  }
  $c \leftarrow c + 1$;
 }
\end{algorithm}

\section{Instance Segmentation Post-Processing}
\label{appendix:postprocessing}

For the instance segmentation experiments we post-process the clustered samples to remove tiny regions that sometimes appear at segmentation borders. For each cluster (instance) found via Algorithm \autoref{algorithm:clustering}, we check whether it survives an erosion operation with an $n \times n$-structure element. If the given erosion eliminates the cluster, we replace each pixel within the cluster in question by its majority neighbour label. The majority neighbour label is determined from a $m \times m$-box centered at the given pixel. The cluster label that is to be replaced as well as background labels are ignored while finding the majority label. If this results in 0 valid neighbour labels, we keep the current pixel's label in SNEMI3D and use the background label in Cityscapes. In both SNEMI3D and Cityscapes, we chose $n=5$ and $m=11$. Painting in the majority label is carried out on the fly.

\section{Model Samples, Reconstructions and (on SNEMI3D and Cityscapes) Instance Clusterings} 
\label{appendix:samples}

\begin{figure}[!h]
\centering
\includegraphics[width=1.5\textwidth, angle=90, origin=c]{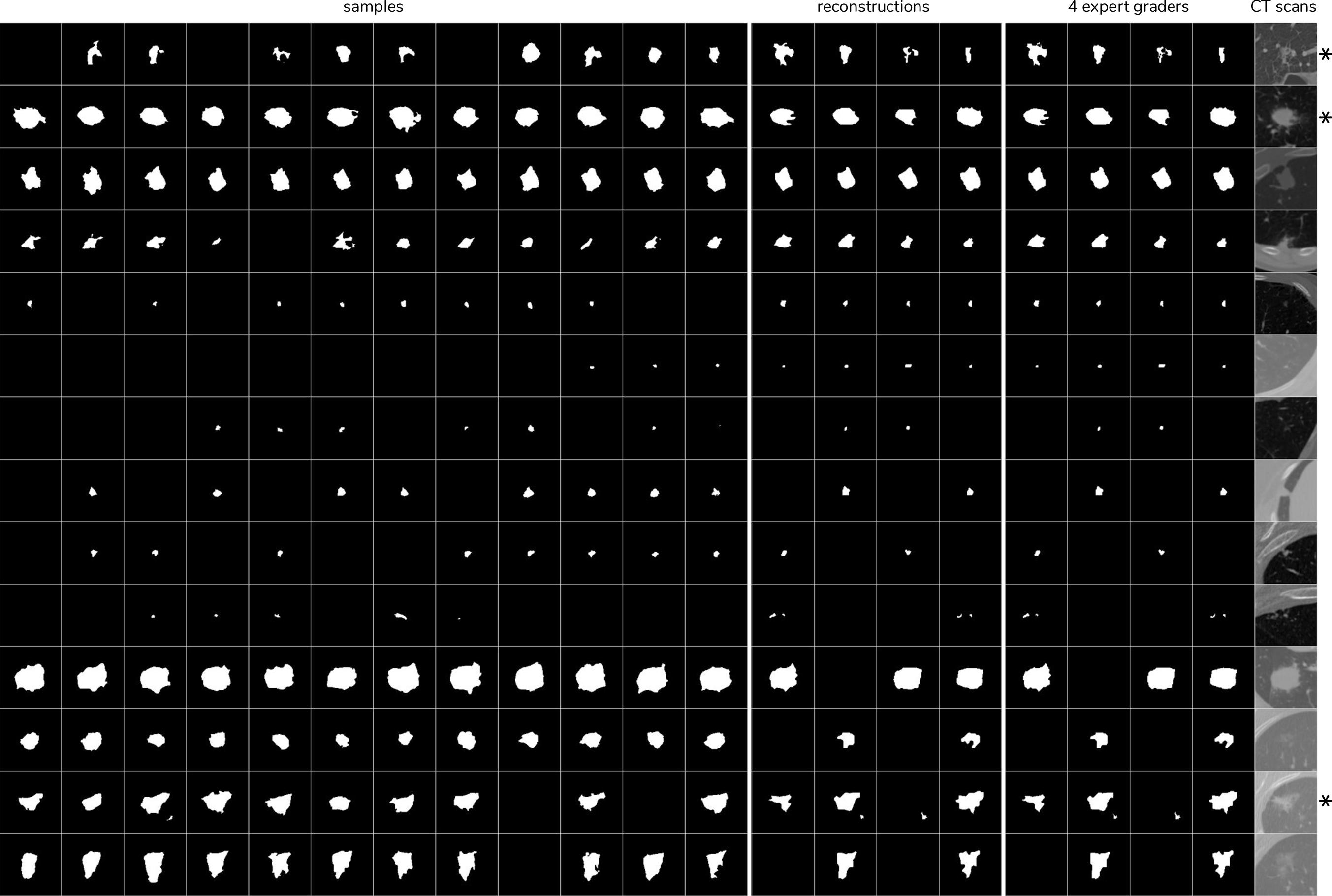}
\caption{Qualitative results of the Hierarchical Probabilistic U-Net on our LIDC-IDRI test set. An * denotes cases that we also use in \autoref{fig:lidc_scales}.}
\label{appendix:hierarch_lidc_samples}
\end{figure}

\begin{figure}[!h]
\centering
\includegraphics[width=1.5\textwidth, angle=90, origin=c]{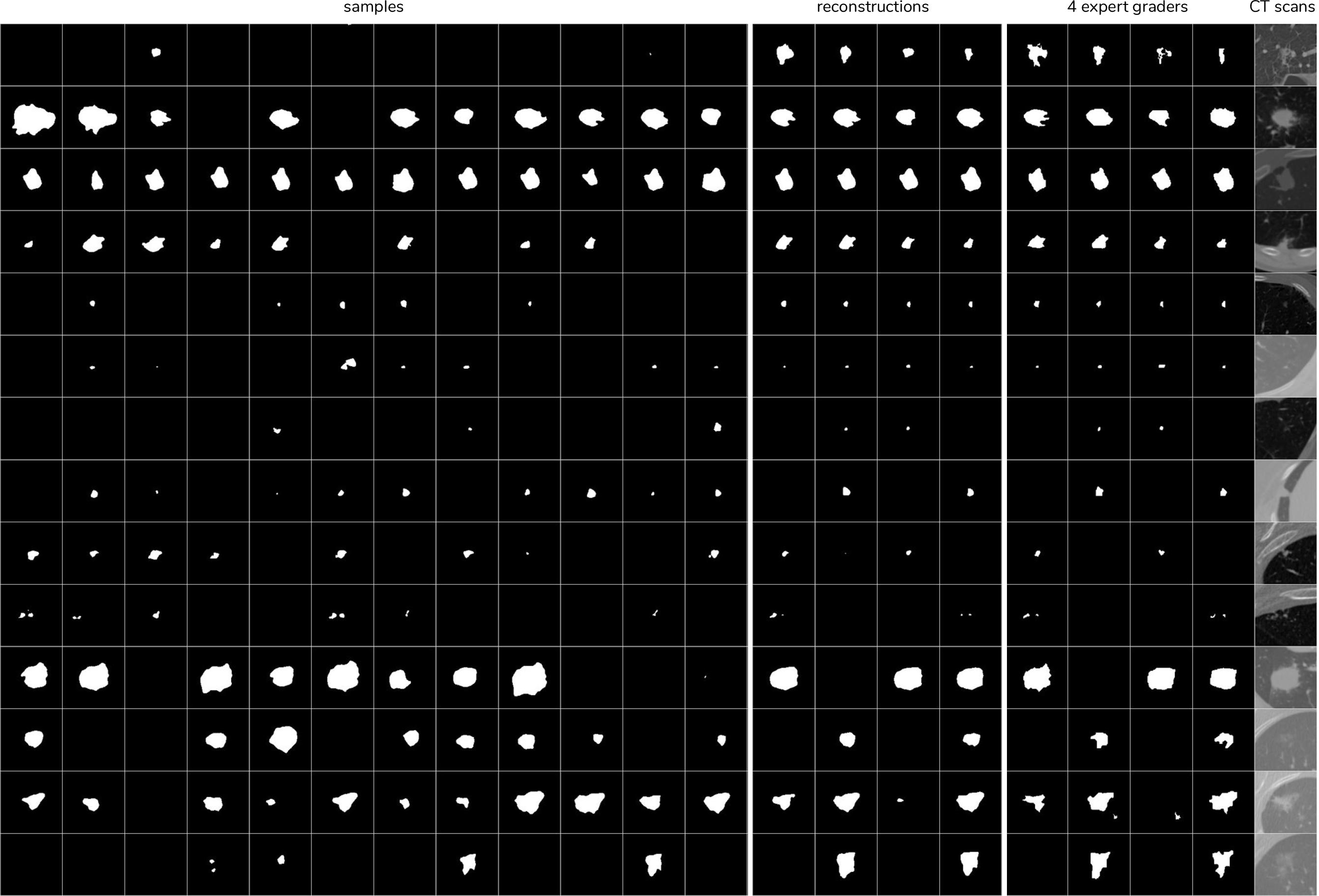}
\caption{Qualitative results of the Standard Probabilistic U-Net on our LIDC-IDRI test set.}
\label{appendix:standard_lidc_samples}
\end{figure}

\begin{figure}[!h]
\centering
\hspace*{-0.5cm}
\includegraphics[width=1.4\textwidth, angle=90, origin=c]{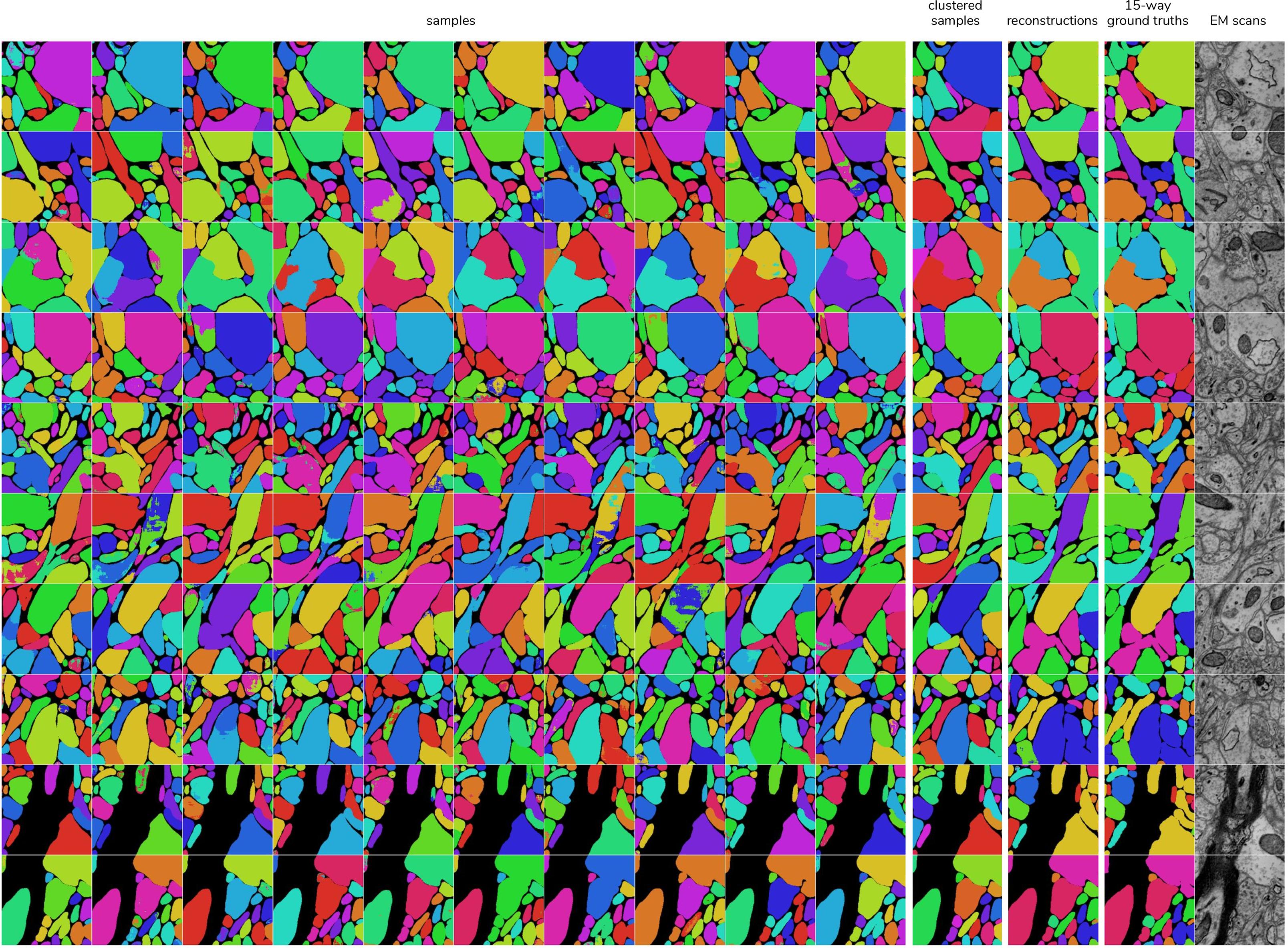}
\caption{Qualitative results of the Hierarchical Probabilistic U-Net on our SNEMI3D test set.}
\label{appendix:hierarch_snemi3d_samples}
\end{figure}

\begin{figure}[!h]
\centering
\hspace*{-0.5cm}
\includegraphics[draft=False,width=1.4\textwidth, angle=90, origin=c]{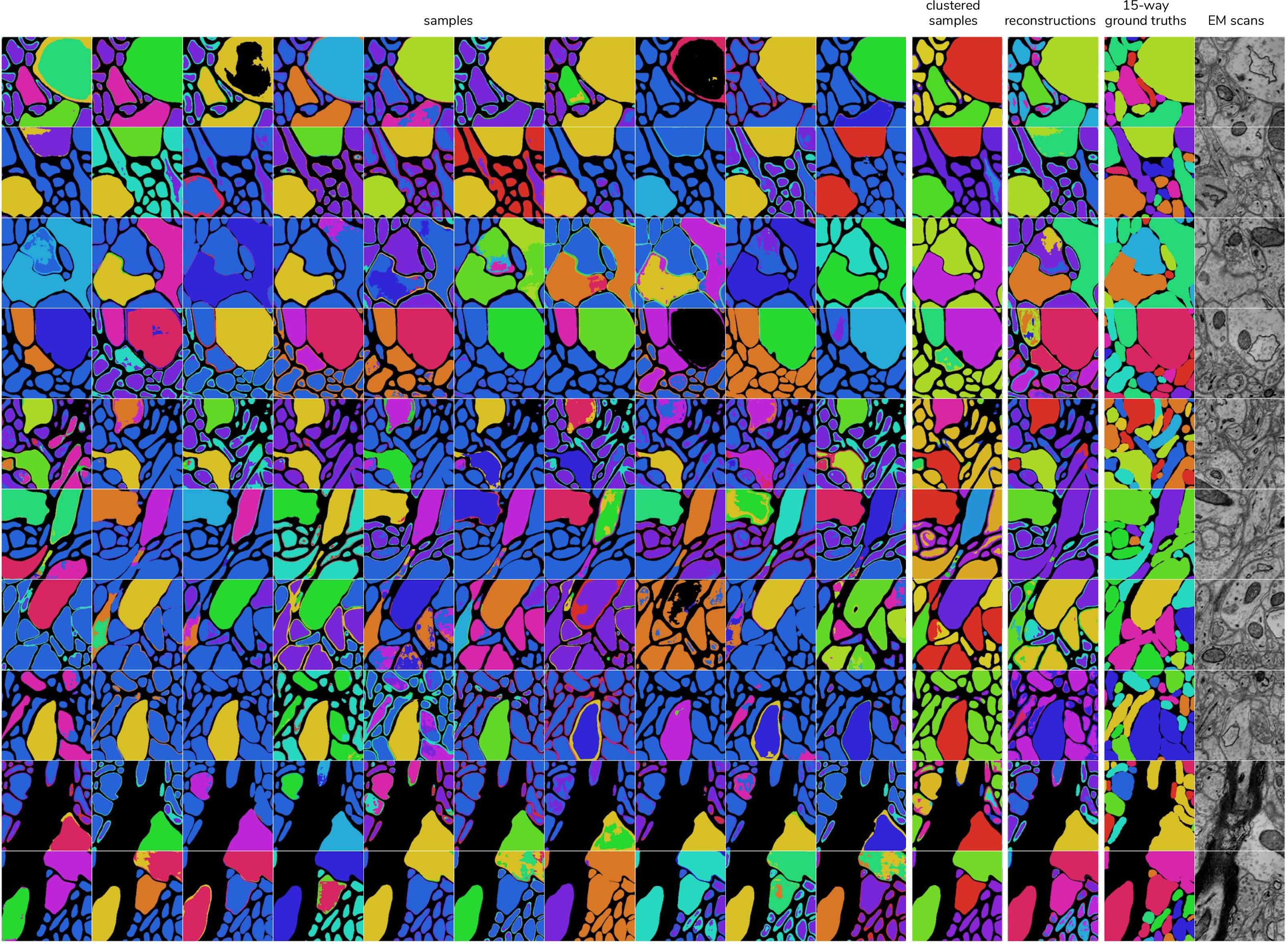}
\caption{Qualitative results of the Standard Probabilistic U-Net on our SNEMI3D test set.}
\label{appendix:standard_snemi3d_samples}
\end{figure}

\begin{figure}[!h]
\centering
\vspace*{-1.6cm}
\hspace*{-1cm}
\includegraphics[draft=False,width=1.65\textwidth, angle=90, origin=c]{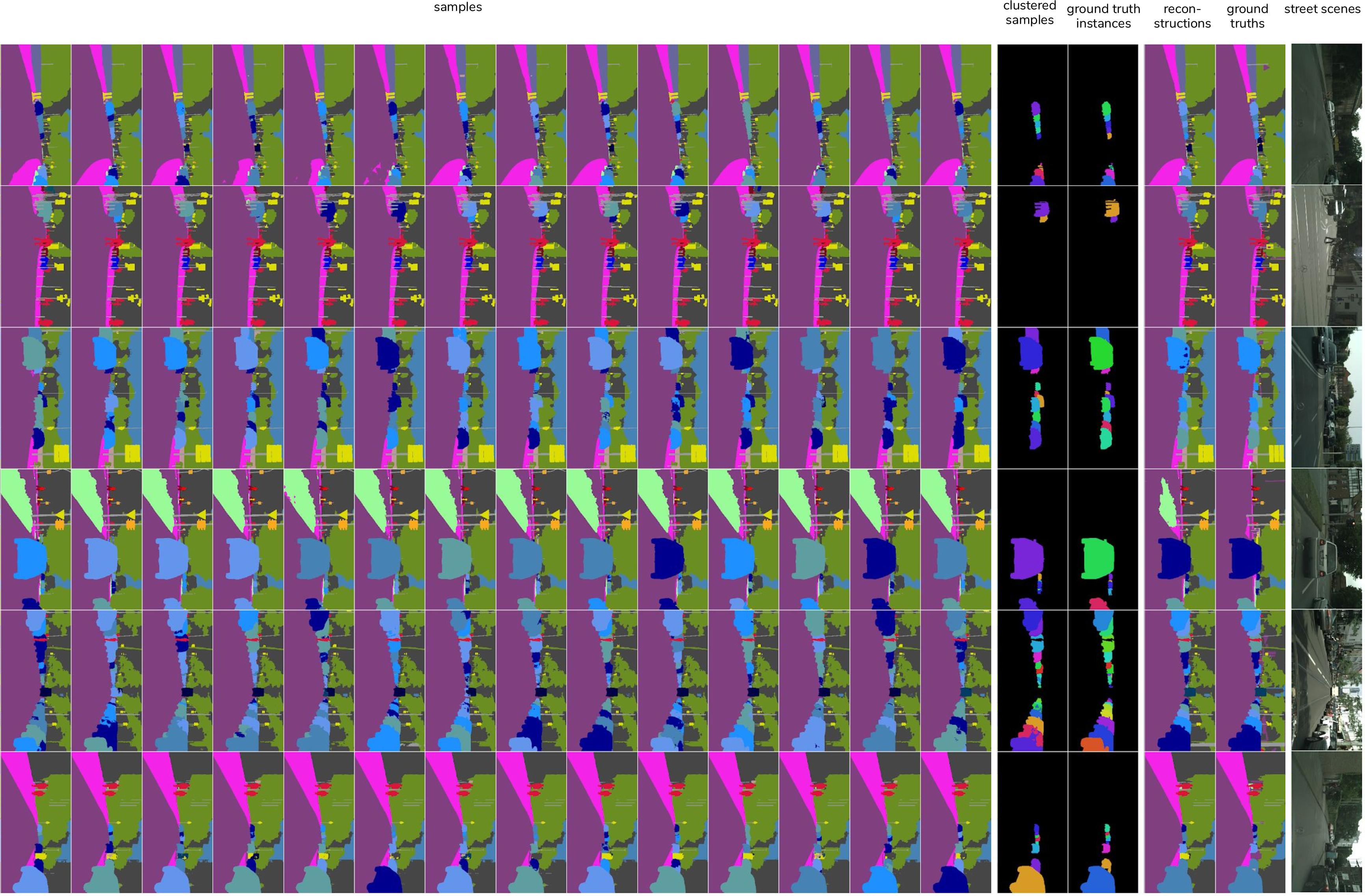}
\caption{Qualitative results of the Hierarchical Probabilistic U-Net on our test set for the Cityscapes car instance task trained with 5 distinct latent car ids on resolution $512 \times 1024$. The 5 car ids take on different shades of blue. The samples show good consistency across individual car instances resulting in high-quality instance segmentations, see the 4$th$ row from the top. Note how the model flips other natural ambiguous regions aside from cars e.g. $\textit{street} \leftrightarrow \textit{sidewalk}$ in the first scene and $\textit{truck} \leftrightarrow \textit{bus}$ in the second last.}
\label{appendix:hierarch_cs_cars_samples}
\end{figure}

\begin{figure}[!h]
\centering
\vspace*{-1.6cm}
\hspace*{-1cm}
\includegraphics[draft=False,width=1.65\textwidth, angle=90, origin=c]{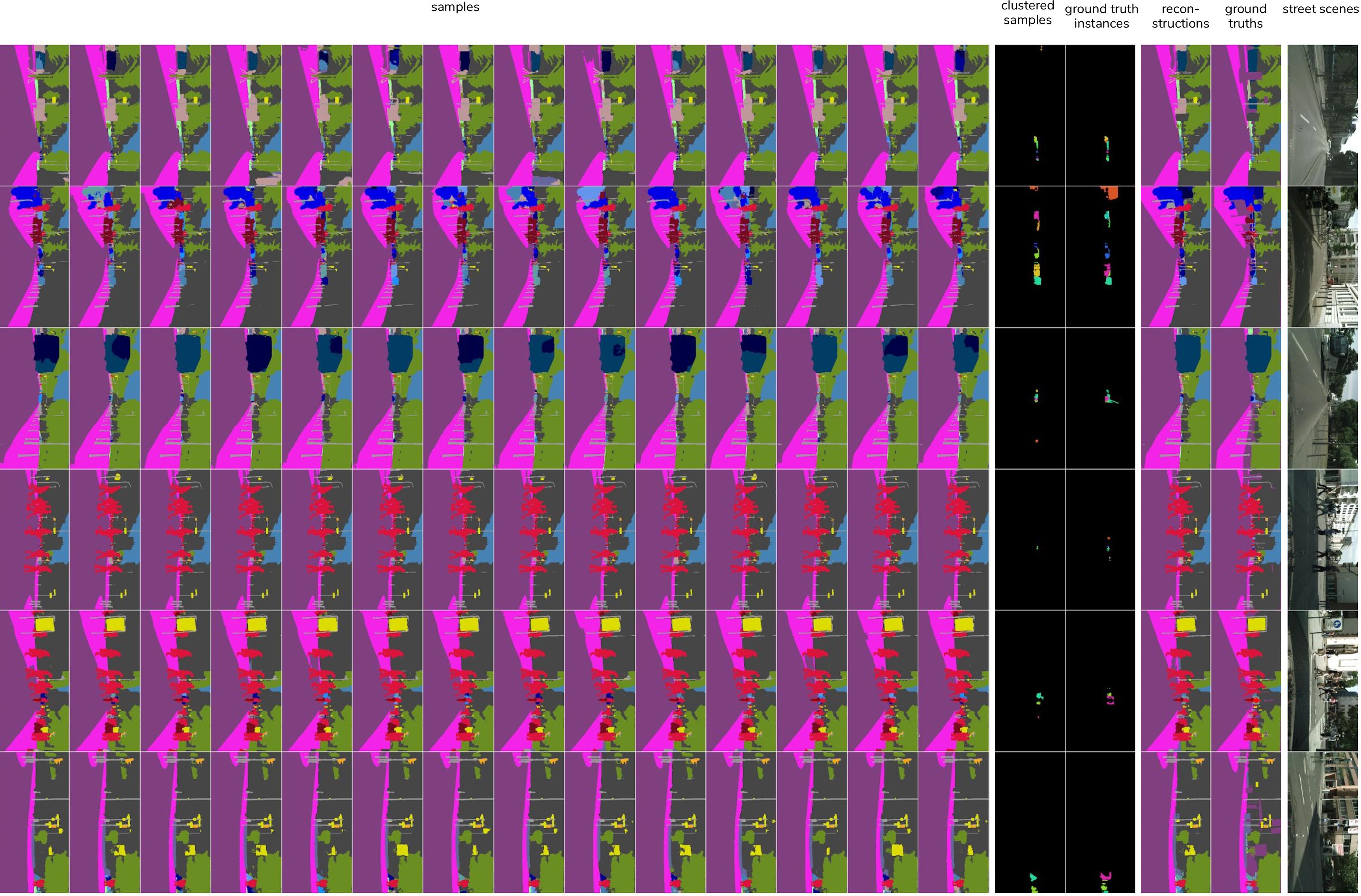}
\caption{Qualitative results of the Hierarchical Probabilistic U-Net on our test set for the Cityscapes car instance task trained with 5 distinct latent car ids on resolution $512 \times 1024$.  The 5 car ids take on different shades of blue. Here we show the top difficult cases in the test set in terms of the Rand error, which shows the difficulty of segmenting individual cars when they are very distant in the scene or heavily occluded.}
\label{appendix:hierarch_cs_cars_samples_difficult}
\end{figure}

\begin{figure}[!h]
\centering
\vspace*{-1.9cm}
\hspace*{-1cm}
\includegraphics[width=1.65\textwidth, angle=90, origin=c]{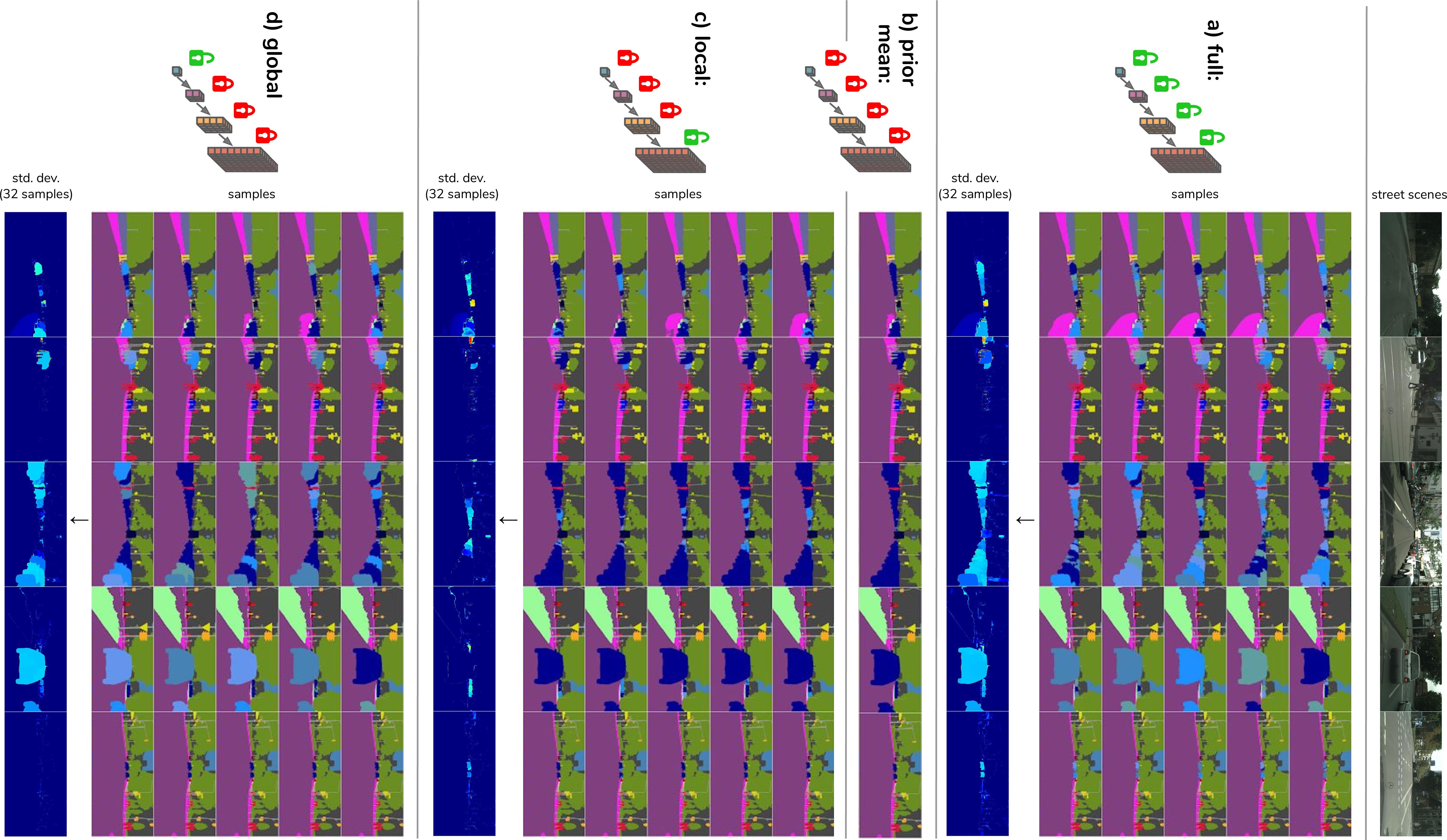}
\caption{Qualitative results of the Hierarchical Probabilistic U-Net on our test set for the Cityscapes car instance task trained with 5 distinct latent car ids on resolution $512 \times 1024$. (\textbf{a}) Samples and standard deviation (std. dev.) across 32 samples when sampling from the full hierarchy. (\textbf{b}) Predictions from the prior mean. (\textbf{c}) Samples and std. dev. when sampling only from the most local scale. (\textbf{d}) Samples and std. dev. when sampling only from the most global scale. Note how the global and local scales affect the instance mask generation almost complementarily.}
\label{appendix:cs_cars_hierarchy}
\end{figure}

\end{document}